\pdfoutput=1

\documentclass[11pt]{article}
\usepackage{amsmath}
\usepackage{colortbl}
\usepackage{svg}

\usepackage[preprint]{acl}

\usepackage{times}
\usepackage{latexsym}

\usepackage[T1]{fontenc}

\usepackage[utf8]{inputenc}

\usepackage{microtype}

\usepackage{inconsolata}

\usepackage{graphicx}

\makeatletter
\def\thanks#1{\protected@xdef\@thanks{\@thanks
        \protect\footnotetext{#1}}}
\makeatother

\title{Multilingual Brain Surgeon: Large Language Models Can be Compressed Leaving No Language Behind}

\author{Hongchuan Zeng$^1$, Hongshen Xu$^1$, Lu Chen$^{1,2{\dagger}}$, Kai Yu$^{1,2{\dagger}}$ \thanks{$\dagger$Lu Chen and Kai Yu are the corresponding authors.}\\
$^1$X-LANCE Lab, Department of Computer Science and Engineering \\ MoE Key Lab of Artificial Intelligence, SJTU AI Institute \\Shanghai Jiao Tong University, Shanghai, China \\ $^2$Suzhou Laboratory, Suzhou, China \\
         \texttt{\{charlie68, xuhongshen, chenlusz, kai.yu\}@sjtu.edu.cn}\\}

\begin{document}
\maketitle

\begin{abstract}
    Large Language Models (LLMs) have ushered in a new era in Natural Language Processing, but their massive size demands effective compression techniques for practicality. Although numerous model compression techniques have been investigated, they typically rely on a calibration set that overlooks the multilingual context and results in significant accuracy degradation for low-resource languages. This paper introduces Multilingual Brain Surgeon (MBS), a novel calibration data sampling method for multilingual LLMs compression. MBS overcomes the English-centric limitations of existing methods by sampling calibration data from various languages proportionally to the language distribution of the model training datasets. Our experiments, conducted on the BLOOM multilingual LLM, demonstrate that MBS improves the performance of existing English-centric compression methods, especially for low-resource languages. We also uncover the dynamics of language interaction during compression, revealing that the larger the proportion of a language in the training set and the more similar the language is to the calibration language, the better performance the language retains after compression. In conclusion, MBS presents an innovative approach to compressing multilingual LLMs, addressing the performance disparities and improving the language inclusivity of existing compression techniques. The codes are available at: \href{https://github.com/X-LANCE/MBS}{https://github.com/X-LANCE/MBS}.
\end{abstract}

\begin{figure*}[htbp]
\centering
\includegraphics[width=\textwidth]{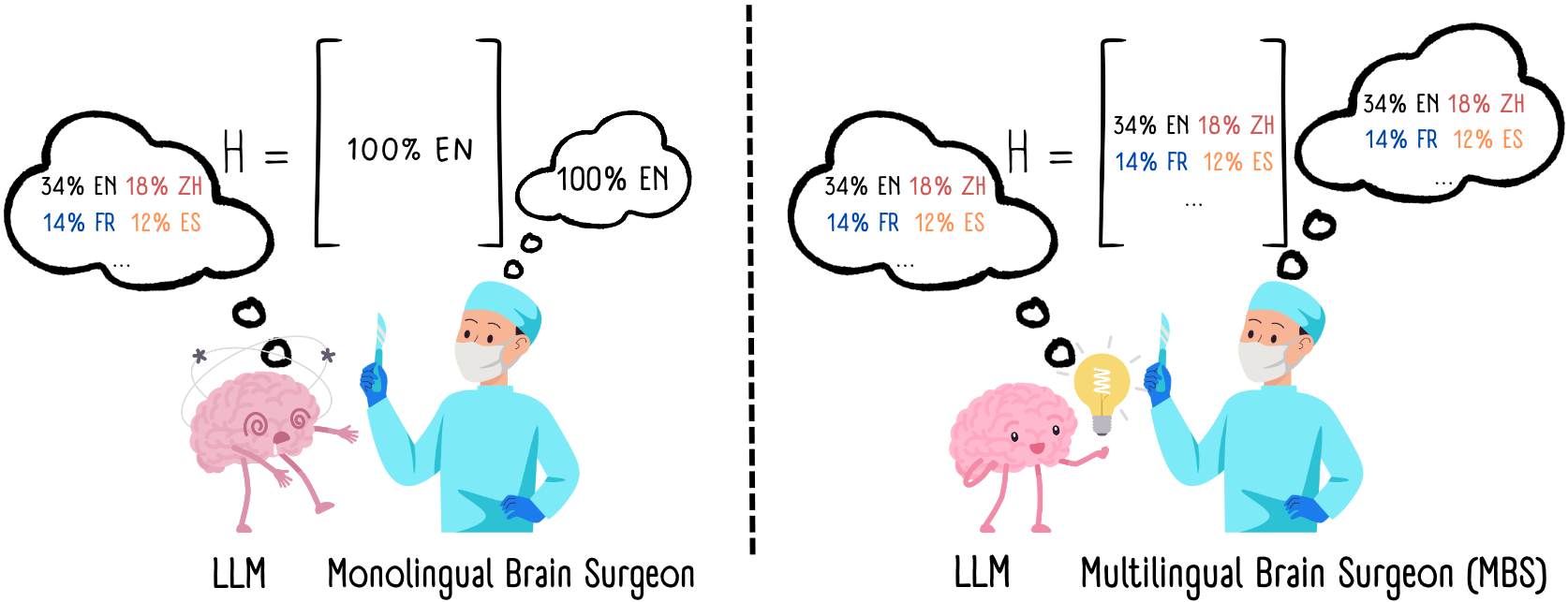}
\caption{MBS samples calibration data from different languages proportionally to the language distribution of training datasets. This approach (right part) effectively addresses the multilingual compression problem compared to previous monolingual sampling methods (left part).}
\label{fig:mbs}
\end{figure*}

\section{Introduction}

Large Language Models (LLMs) have revolutionized Natural Language Processing (NLP) with their remarkable performance. However, their colossal size and computational demands necessitate effective Model Compression (MC) techniques for practical use. In the case of multilingual LLMs, the vast size is crucial for retaining information from various languages and mitigating the curse of multilinguality \citep{conneau2020unsupervised,goyal2021larger}. Moreover, wide language coverage and interference among languages pose a harder challenge for compressing multilingual LLMs.

Existing approaches for MC have predominantly focused on model quantization \cite{frantar2023gptq,dettmers2022llmint8,xiao2023smoothquant,yao2022zeroquant}, where model parameters are mapped to lower bit-level representations, and network pruning, which reduces the size of neural networks by eliminating unnecessary connections. Inspired by the classic Optimal Brain Damage (OBD) and Optimal Brain Surgeon (OBS) pruning framework \cite{298572,10.5555/2969830.2969903}, various approaches, namely GPTQ \citep{frantar2023gptq} for model quantization, SparseGPT \citep{frantar2023sparsegpt} and Wanda \citep{sun2023simple} for network pruning, have been proposed to compress  LLMs. These compression methods utilize a calibration dataset to determine the priority of parameters and thus are retraining-free, avoiding expensive fine-tuning cost especially for LLMs.

However, neither of these methods has considered the multilingual scenario: all of them use a single-language (e.g., English) calibration dataset to determine the priority of parameters for multilingual models. A significant performance drop on multilingual tasks is observed due to this English-centric approach, especially in the case of low-resource languages.

In this paper, we propose Multilingual Brain Surgeon (MBS), which has successfully achieved significant sparsity levels when compressing multilingual LLMs while simultaneously minimizing the performance drop across different languages in the models, leaving no language behind after compression. Specifically, as shown in Figure \ref{fig:mbs}, MBS samples the calibration data of different languages proportionally to the language distribution of the model training dataset. This approach effectively addresses the multilingual compression problem compared to previous monolingual sampling methods. Furthermore, we observed the dynamics of language interaction during compression and drew two main conclusions: 1) \textit{The larger the proportion of a language in the model training dataset}, the more resistant it is to compression. 2) \textit{The more similar the downstream language is to the calibration language}, the less performance drop it obtained after compression. We further propose a measure of similarity among languages to explain and predict the performance drop.

The experiments were conducted on BLOOM \citep{bigscience_workshop_2022}, one of the most effective open-source multilingual LLM models. We sample the calibration data from CC-100 \citep{wenzek-etal-2020-ccnet}, a widely used dataset of web-crawled data containing 100+ languages. The perplexity of languages is tested on XL-Sum \citep{hasan-etal-2021-xl}, a dataset that contains high-quality articles from BBC covering 45 languages. Experimental results demonstrate that MBS enhances the performance of GPTQ, SparseGPT, and Wanda compared to using only English calibration data. We want to further highlight that MBS is applicable to all compression methods that involve the use of calibration data, especially those following the OBS/OBD framework \citep{298572, 10.5555/2969830.2969903}, which necessitates approximations of second-derivative information.

\section{Background}

\subsection{Optimal Brain Surgeon (OBS)}

Optimal Brain Surgeon \citep{298572} is a classic network pruning algorithm. It assumes that a network's error converges to a local minimum and calculates the second-order derivatives (Hessian matrix $\mathbf{H}$) of the error ($E$) with respect to each parameter ($w$) to determine which connections can be safely pruned without significantly affecting performance. The increase in error ($L_j$) when a parameter ($w_j$) is set to zero, and the optimal adjustment ($\delta w$) of the remaining weights to compensate for the removal are given by:
\begin{equation}\label{eq:1}
L_j = \frac{1}{2} \frac{{w_j}^2}{[\mathbf{H}^{-1}]_{jj}}
\end{equation}
\begin{equation}
\delta w = -\frac{{w_j}}{[\mathbf{H}^{-1}]_{jj}}\mathbf{H}^{-1}_{:,j}.
\end{equation}

\subsection{Error Measurement}

The network's error can be expressed in terms of the $l_2$-error between the outputs before and after compression \citep{hubara2021accelerated}. Given inputs $\mathbf{X}$ (the training dataset), the original weights $\mathbf{W}$, the updated weights $\mathbf{\hat{W}}$, and a sparsity mask $\mathbf{M}$ of the same size as $\mathbf{W}$, the error is defined as:
\begin{equation}
E = ||\mathbf{WX - (M \odot \hat{W})X}||_2^2.
\end{equation}
In the case of quantization, the mask is a matrix filled with ones. The second-order derivatives ($\mathbf{H}$) of the error with respect to the parameters are therefore represented as $\mathbf{H = 2XX^T}$, which forms the basis of our approximation objective.

\subsection{SparseGPT, Wanda and GPTQ}\label{SparseGPT and Wanda}

To assess the importance of parameters, SparseGPT and Wanda employ different pruning metrics. Taking inspiration from OBS, SparseGPT defines its metric as $S_{i,j}=[|\mathbf{W}|^2/\mathbf{diag}((\mathbf{X^TX}+\lambda \mathbf{I})^{-1})]_{i,j}$, with $\lambda$ being the Hessian dampening factor to prevent inverse computation collapse. On the other hand, Wanda uses $S_{i,j}=|\mathbf{W}_{i,j}|\cdot ||\mathbf{X}_j||_2$ as its pruning metric.

Remarkably, these two metrics are essentially equivalent when $\lambda$ is set to 0, and only the diagonal elements of the Hessian matrix $X^TX+\lambda I$ are retained:
\begin{equation}
\mathbf{diag}(((\mathbf{X^TX+\lambda I)\odot I})^{-1}) =(||\mathbf{X}_j||^2_2)^{-1}.
\end{equation}
This assumption aligns with the practice of Optimal Brain Damage \citep{10.5555/2969830.2969903}, which retains only the diagonal elements of the second-order derivatives matrix. Consequently, we can conclude that:
\begin{equation}
\mathbf{S}_{SparseGPT} = \mathbf{S}_{Wanda}^2
\end{equation}
if we disregard the non-diagonal elements of $\mathbf{H}$.

The primary distinctions between SparseGPT and Wanda are as follows:
\begin{itemize}
\setlength{\itemsep}{0pt}
\setlength{\parsep}{0pt}
\setlength{\parskip}{0pt}
\item SparseGPT retains the non-diagonal elements of the Hessian metrics, whereas Wanda takes the opposite approach.
\item SparseGPT performs adjustments ($\delta w$) on non-pruned parameters to compensate for removal, while Wanda does not.
\end{itemize}
Equally inspired by OBD, the quantization formulas provided by GPTQ are as follows:
\begin{equation}
w_j=\operatorname{\mathbf{argmin}}_{w_j} \frac{\left(\operatorname{\mathbf{quant}}\left(w_j\right)-w_j\right)^2}{\left[\mathbf{H}^{-1}\right]_{j j}}
\end{equation}
\begin{equation}
\boldsymbol{\delta}_F=-\frac{w_j-\operatorname{\mathbf{quant}}\left(w_j\right)}{\left[\mathbf{H}^{-1}\right]_{j j}} \cdot\left(\mathbf{H}^{-1}\right)_{:, j}
\end{equation}
Here, $w_j$ represents the greedy-optimal weight to quantize next, $\boldsymbol{\delta}_F$ denotes the corresponding optimal update of weights, and $\mathbf{quant}(w)$ rounds the value of $w$ to the nearest point on the quantization grid. It's evident that these formulas follow a similar pattern to the OBD/OBS approach, and the information of the Hessian matrix $\mathbf{H}$ is crucial in all these methods.

\section{Is Monolingual Calibrating Applicable to Multilingual MC?}


Previous model compression methods only use English corpus as the sole calibration data, neglecting other languages. This raises the question: \textbf{how does monolingual calibration impact the performance of other languages during multilingual model compression?} In this section, we aim to explore this issue theoretically, focusing on two main aspects: the proportion of languages in the training data, and the similarity between languages. Further experimental analysis will be provided in Section \ref{sec:mmc}.



We denote the total error of the model as $E$, and the error on language $m$ as $E_m$. We know that model training convergence applies to the whole training dataset. Thus, $E$ resides in a local minimum. However, for languages $m$ and $n$, $E_m$ and $E_n$ may not necessarily be in their own local minima. This also explains the presence of the multilingual curse \cite{conneau2020unsupervised}, where the performance of a multilingual model in all languages is lower than that of a monolingual model with the same configuration. This occurs because the model is in a global local minimum, rather than individual local minima for each language. Due to their differing distributions, the local minima for each language do not overlap. The reason why using larger language models can alleviate this problem might be that, with a huge amount of parameters, they can simulate a distribution sophisticated enough where different languages' local minima are close. 

\subsection{Proportion in training data}

Due to the fact that the size of English corpus is much larger than low-resource languages, we may suppose a language pair $m$ and $n$ with a significantly different training corpus size ($p_n>>p_m$). 

Intuitively, we can assume that languages with larger corpora in the training set tend to have their minimum error closer to the minimum of $E$ because they contribute more weight to the total error. This characteristic makes them more robust against compression. Conversely, languages with smaller corpora inherently have their minimum error farther from the minimum of $E$, and compression can potentially push them even further away.

\begin{figure}[h]
    \centering
\includegraphics[width=0.45\textwidth]{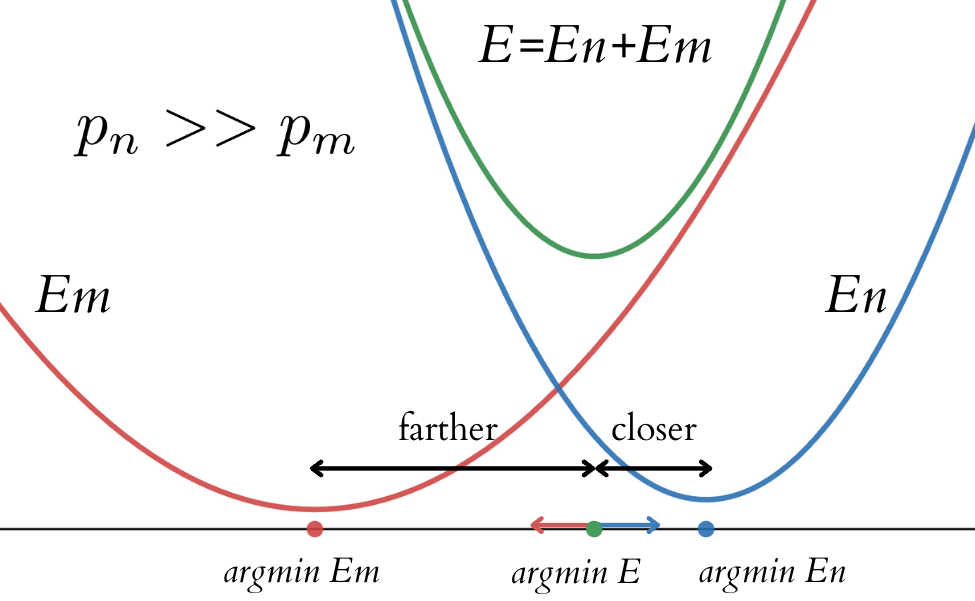}
    \caption{Languages with larger corpora have their minimum error closer to the minimum of $E$. Monolingual compression effectively "pushed" the model's state towards the minimum error of that particular language.}
    \label{dataAmount}
\end{figure}

This phenomenon is manifested in the following way illustrated in Figure \ref{dataAmount}: \textit{when compressing models with only the calibration data of the well-represented}\footnote{In the rest of the paper, we call a language "well-represented" when its proportion is relatively big in the model training set, and "underrepresented" when its proportion is relatively small.} \textit{language $n$, it has a significant impact on the performance of the underrepresented language $m$. However, compressing models with only the calibration data of the underrepresented language $m$ has a comparatively minor impact on the performance of the well-represented language $n$}.

\subsection{Similarity between languages}

In the second scenario, we may suppose that the two languages are as well-represented as each other ($p_m \approx p_n$). According to Equation \ref{eq:1}, the priority of compression is fully determined by $\mathbf{H}$, so it is sufficient to compare $\mathbf{H_m}$ and $\mathbf{H_n}$. We may suppose the non-diagonal elements are trivial \citep{10.5555/2969830.2969903} to calculate the inverse of $\mathbf{H}$. The metric is thus simplified to $S = \mathbf{|W| \cdot ||X||_2}$, so we can directly compare $||\mathbf{X}||_2$, which is a vector of length $q$ (number of parameters), and each of the elements is the sum of the square of the inputs at the corresponding position.

\begin{figure}[htb!]
    \centering
    \includegraphics[width=0.5\textwidth]{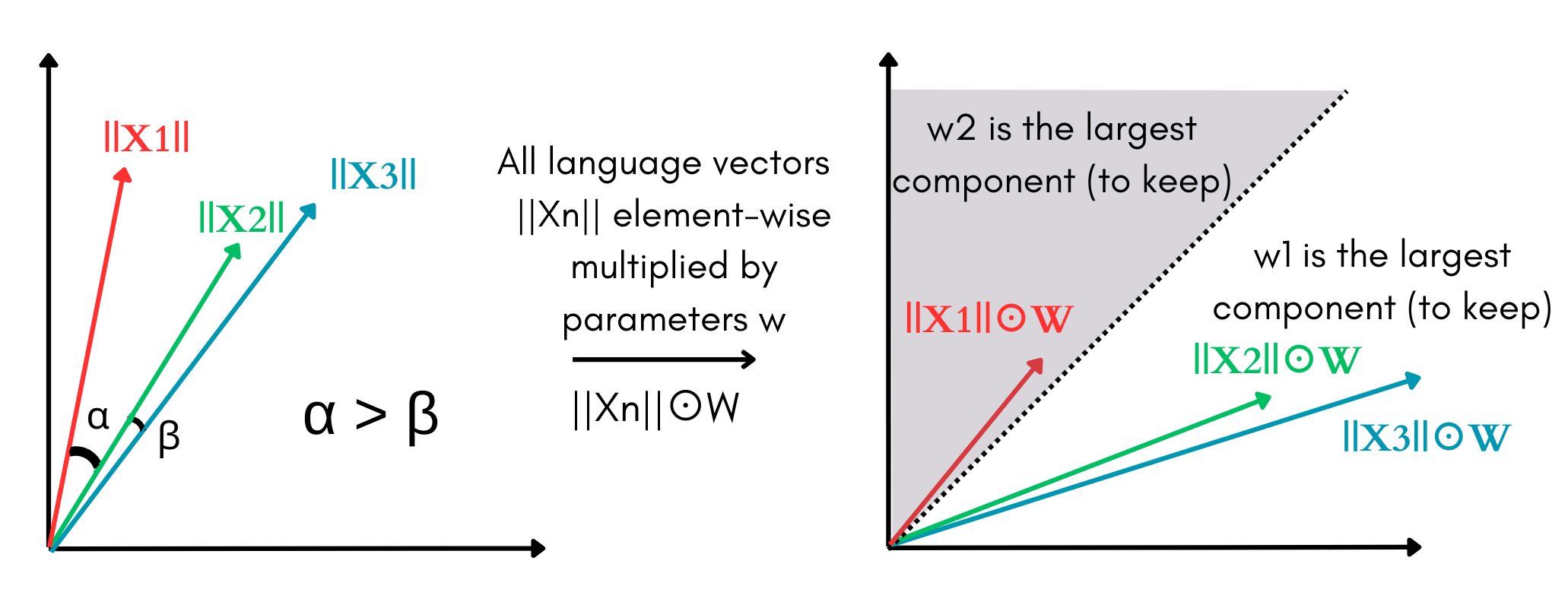}
    \caption{The angle between language 2 and language 3 is smaller than that between language 1 and language 2. After element-wise multiplication, language 2 and 3 are more likely to prioritize the same parameter $w_1$ because their angle before multiplication is smaller.}
    \label{whycos}
\end{figure}

A classic method to compare the similarity of two vectors is cosine similarity. The choice of cosine similarity over Euclidean distance is motivated by the need to compare two vectors based on the likelihood that their largest components remain consistent after undergoing the same element-wise multiplication with unknown vectors (model parameters). This can be modeled as the comparison of two vectors after they have experienced the same coordinate axis transformation, assessing whether their largest components remain identical. Clearly, when two vectors have a smaller angle between them, the likelihood that their largest components remain the same after undergoing the same coordinate axis transformation is relatively higher (demonstrated in Figure \ref{whycos}). 

However, it's important to acknowledge that cosine similarity does not fulfill the properties of a distance metric, particularly the triangle inequality. Consequently, we cannot directly deduce the similarity between languages 1 and 3 from the similarities between 1 and 2, and 2 and 3. However, the property of a distance metric is less critical in the context of our work, since our goal is only to compare the similarity between the calibration language and the non-calibration languages, rather than among non-calibration languages.

We can compute the cosine similarity between $||\mathbf{X}_m||_2$ and $||\mathbf{X}_n||_2$. When they are similar, using only data of language $m$ as calibration data will introduce little performance drop in language $n$, and vice versa. \textit{That is to say, when two languages are very different, employing data from just one of the two languages as calibration data will lead to a significant performance decrease in the other.}

\section{Multilingual Brain Surgeon (MBS)}
To mitigate interference among languages in multilingual model compression, we introduce Multilingual Brain Surgeon (MBS), a method that proportionally samples calibration data from different languages based on their distribution in the model training dataset. We provide additional theoretical details as follows.

In the OBD/OBS framework, we treat the error ($E$) as a whole. This makes sense for monolingual models since they contain only one language. However, for multilingual models, the error can be regarded as the sum of errors ($E_n$) associated with different languages. For a model trained on multiple languages, we can express the total error as follows:
\begin{equation}
E = E_1 + E_2 + E_3 + \ldots + E_n.
\end{equation}

Consequently, the Hessian matrix can be represented as the sum of Hessian matrices for each language:
\begin{equation}
\mathbf{H = H_1 + H_2 + H_3 + \ldots + H_n},
\end{equation}
where $\mathbf{H_n = {X_n}^TX_n}$. Here, $\mathbf{X_n}$ represents the inputs (training data) for language $n$, with a shape of $q\times p_n$, where $q$ is the total number of network parameters, and $p_n$ is the total number of training samples for language $n$.

Let's denote a subset of training data as $X_n^{[k]}$. Then, we have:
\begin{equation}
\mathbf{H_n} = \mathbf{{X_n}^TX_n} = \sum_{k=1}^{p_n} {X_n^{[k]}}^TX_n^{[k]},
\end{equation}
which leads to:
\begin{equation}
\begin{split}
\mathbf{H} = \sum_{k=1}^{p_1} {X_1^{[k]}}^TX_1^{[k]} + \sum_{k=1}^{p_2} {X_2^{[k]}}^TX_2^{[k]} \\+ \ldots + \sum_{k=1}^{p_n} {X_n^{[k]}}^TX_n^{[k]}.
\end{split}
\end{equation}

It's evident that each language's contribution to $\mathbf{H}$ depends on its representation in the model's training data. Therefore, when selecting calibration data, it's essential to choose samples from each language in proportion to its presence in the training set. Specifically, for language $n$, the percentage of its representation in the training set is $p_n/p$, where $p$ is the total number of training samples. \textit{Thus, we should allocate a proportionate amount of data from language $n$ (i.e., $p_n/p$ percent) in the calibration data used for compression.}

\section{Experiments}

\subsection{Experimental Setup}
\begin{figure*}[!htb]
    \centering
    \includegraphics[width=\textwidth]{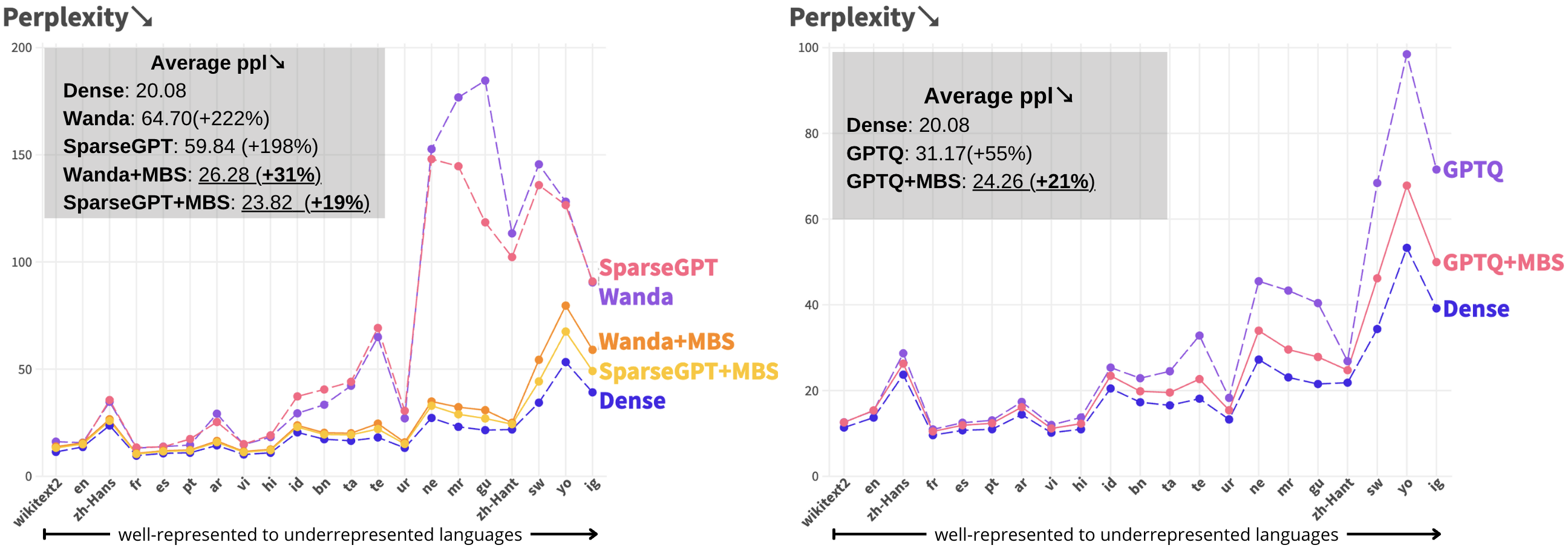}
    \caption{Perplexity for each language and their respective increases when compared to the dense BLOOM-7b1 model after pruning (left) or quantization (right). From left to right, languages are ranked in order from the most well-represented to the least represented.}
    \label{ppl_mono_7b1}
\end{figure*}

\textbf{Models.} The experiments were conducted using the BLOOM \citep{bigscience_workshop_2022} model family, which is recognized as one of the most effective open-source multilingual LLMs. Our primary tests were performed on both the BLOOM-560m and BLOOM-7b1 models to provide insights into the performance of smaller and larger models. For the network pruning experiments, a pruning sparsity of 50\% was applied. In the quantization experiments, the models were quantized to 3 bits precision with groupings of size 1024.

\textbf{Datasets \& Language Selection.} For calibration data, we selected CC-100 \citep{wenzek-etal-2020-ccnet}, a dataset comprising web-crawled content in over 100 languages, similar to the setup used by previous studies like \citet{frantar2023sparsegpt}, \citet{sun2023simple}, and \citet{frantar2023gptq} which used a monolingual English dataset called C4 \citep{2019t5}.

To evaluate multilingual perplexity, we employed XL-Sum \citep{hasan-etal-2021-xl}, a dataset containing high-quality articles from BBC covering 45 languages, as our benchmark. Additionally, we assessed perplexity on the test sets of raw-WikiText2 \citep{merity2016pointer}, a widely used English perplexity benchmark. Due to resource limitations for certain languages in the BLOOM model, we conducted experiments on a subset of 20 languages, which were those available in CC-100, XL-Sum, and BLOOM. These languages include Arabic (\texttt{ar}), Bengali (\texttt{bn}), Chinese simplified (\texttt{zh-Hans}), Chinese traditional (\texttt{zh-Hant}), French (\texttt{fr}), Gujarati (\texttt{gu}), Hindi (\texttt{hi}), Igbo (\texttt{ig}), Indonesian (\texttt{id}), Marathi (\texttt{mr}), Nepali (\texttt{ne}), Portuguese (\texttt{pt}), Spanish (\texttt{es}), Swahili (\texttt{sw}), Tamil (\texttt{ta}), Telugu (\texttt{te}), Urdu (\texttt{ur}), Vietnamese (\texttt{vi}), and Yoruba (\texttt{yo}).

\textbf{Evaluation.} We evaluated the perplexity of the compressed model separately for each language using XL-Sum. We also conducted zero-shot evaluations, employing the widely recognized EleutherAI-eval-harness \citep{eval-harness}, with a focus on multilingual tasks to assess the performance of less-represented languages. The zero-shot tasks that we have chosen to evaluate the compressed model are specified in Table \ref{0-shot7b1}.

\textbf{Calibration data \& Baselines.} Our calibration data consisted of 256 segments, each containing 2048 tokens, sampled from CC-100. We used our MBS sampling method and sampled 87, 47, 37, 31, 14, 13, 7, 4, 3, 3, 1, 1, 1, 1, 1, 1, 1, 1, 1, 1 segments respectively for \texttt{en}, \texttt{zh-Hans}, \texttt{fr}, \texttt{es}, \texttt{pt}, \texttt{ar}, \texttt{vi}, \texttt{hi}, \texttt{id}, \texttt{bn}, \texttt{ta}, \texttt{te}, \texttt{ur}, \texttt{ne}, \texttt{mr}, \texttt{gu}, \texttt{zh-Hant}, \texttt{sw}, \texttt{yo} and \texttt{ig}. Additionally, we conducted tests in the \textit{Equal MBS} setting, in which an equal number of segments were sampled from each language. We also implemented the monolingual compression setting, using 256 segments from the same language.

\textbf{Language Similarity Study.} To study language similarity, we conducted monolingual pruning on English and Igbo, representing the best-represented and worst-represented languages in our dataset, respectively. We also performed similar experiments on Urdu and Tamil, which respectively represent the least and most similar languages to the others (further explanation is provided in the results section). To compare language similarity, we utilized the representations after the embedding layer of the BLOOM model, as the compression algorithms do not affect the embedding layer.

\subsection{Main results}

\begin{table*}
\small
\centering
\begin{tabular}{l
>{\columncolor[HTML]{F2F2F2}}c c
>{\columncolor[HTML]{D9D9D9}}c c
>{\columncolor[HTML]{D9D9D9}}c c
>{\columncolor[HTML]{D9D9D9}}c }
\textbf{\begin{tabular}[c]{@{}l@{}}Accuracy of \\ 0-shot task\end{tabular}} & \textbf{Dense}                    & \textbf{Wanda}                    & \textbf{\begin{tabular}[c]{@{}r@{}}Wanda\\ +MBS\end{tabular}} & \textbf{\scriptsize {SparseGPT}}                & \textbf{\begin{tabular}[c]{@{}r@{}}\scriptsize SparseGPT\\ \scriptsize +MBS\end{tabular}} & \textbf{GPTQ}                     & \textbf{\begin{tabular}[c]{@{}r@{}}GPTQ\\ +MBS\end{tabular}} \\
\cellcolor[HTML]{BFBFBF}\textbf{xcopa$\uparrow$}                                     & \cellcolor[HTML]{BFBFBF}\textbf{} & \cellcolor[HTML]{BFBFBF}\textbf{} & \cellcolor[HTML]{BFBFBF}\textbf{}                             & \cellcolor[HTML]{BFBFBF}\textbf{} & \cellcolor[HTML]{BFBFBF}\textbf{}                                 & \cellcolor[HTML]{BFBFBF}\textbf{} & \cellcolor[HTML]{BFBFBF}\textbf{}                            \\
\textbf{id}                                                                 & 69.80\%                           & 67.20\%                           & \textit{\textbf{67.40\%}}                                     & 65.60\%                           & \textit{\textbf{66.40\%}}                                         & 67.20\%                           & \textit{\textbf{67.40\%}}                                    \\
\textbf{sw}                                                                 & 51.60\%                           & 54.80\%                           & 53.80\%                                                       & 55.20\%                           & 51.20\%                                                           & 54.60\%                           & \textit{\textbf{55.00\%}}                                    \\
\textbf{ta}                                                                 & 59.20\%                           & 61.20\%                           & 57.80\%                                                       & 60.60\%                           & 58.60\%                                                           & 58.40\%                           & 57.80\%                                                      \\
\textbf{vi}                                                                 & 70.80\%                           & 69.80\%                           & 67.20\%                                                       & 66.80\%                           & 66.40\%                                                           & 67.00\%                           & \textit{\textbf{68.20\%}}                                    \\
\textbf{zh}                                                                 & 65.20\%                           & 62.00\%                           & \textit{\textbf{63.60\%}}                                     & 62.20\%                           & \textit{\textbf{63.80\%}}                                         & 61.00\%                           & \textit{\textbf{62.60\%}}                                    \\
\textbf{Average}                                                            & 63.32\%                           & 63.00\%                           & 61.96\%                                                       & 62.08\%                           & 61.28\%                                                           & 61.64\%                           & \textit{\textbf{62.20\%}}                                    \\
\cellcolor[HTML]{BFBFBF}\textbf{xstory\_cloze$\uparrow$}                             & \cellcolor[HTML]{BFBFBF}          & \cellcolor[HTML]{BFBFBF}          & \cellcolor[HTML]{BFBFBF}\textit{\textbf{}}                    & \cellcolor[HTML]{BFBFBF}          & \cellcolor[HTML]{BFBFBF}\textit{\textbf{}}                        & \cellcolor[HTML]{BFBFBF}          & \cellcolor[HTML]{BFBFBF}\textit{\textbf{}}                   \\
\textbf{ar}                                                                 & 58.57\%                           & 53.94\%                           & \textit{\textbf{54.93\%}}                                     & 54.93\%                           & \textit{\textbf{56.32\%}}                                         & 56.45\%                           & \textit{\textbf{57.18\%}}                                    \\
\textbf{en}                                                                 & 70.75\%                           & 68.23\%                           & 67.70\%                                                       & 69.23\%                           & 68.96\%                                                           & 68.70\%                           & \textit{\textbf{68.96\%}}                                    \\
\textbf{es}                                                                 & 66.12\%                           & 64.39\%                           & 63.20\%                                                       & 62.87\%                           & \textit{\textbf{64.39\%}}                                         & 64.53\%                           & \textit{\textbf{64.79\%}}                                    \\
\textbf{hi}                                                                 & 60.56\%                           & 56.92\%                           & \textit{\textbf{57.18\%}}                                     & 57.64\%                           & \textit{\textbf{58.44\%}}                                         & 58.04\%                           & \textit{\textbf{58.17\%}}                                    \\
\textbf{id}                                                                 & 64.46\%                           & 59.96\%                           & \textit{\textbf{60.29\%}}                                     & 59.23\%                           & \textit{\textbf{61.81\%}}                                         & 60.89\%                           & \textit{\textbf{62.54\%}}                                    \\
\textbf{sw}                                                                 & 53.94\%                           & 50.89\%                           & \textit{\textbf{51.69\%}}                                     & 50.69\%                           & \textit{\textbf{52.02\%}}                                         & 52.28\%                           & \textit{\textbf{52.95\%}}                                    \\
\textbf{te}                                                                 & 57.45\%                           & 56.52\%                           & \textit{\textbf{56.72\%}}                                     & 56.78\%                           & \textit{\textbf{57.97\%}}                                         & 57.18\%                           & \textit{\textbf{57.71\%}}                                    \\
\textbf{zh}                                                                 & 61.88\%                           & 58.37\%                           & \textit{\textbf{59.56\%}}                                     & 57.91\%                           & \textit{\textbf{60.89\%}}                                         & 60.03\%                           & \textit{\textbf{60.03\%}}                                    \\
\textbf{Average}                                                            & 61.71\%                           & 58.65\%                           & \textit{\textbf{58.91\%}}                                     & 58.66\%                           & \textit{\textbf{60.10\%}}                                         & 59.76\%                           & \textit{\textbf{60.29\%}}                                    \\
\cellcolor[HTML]{BFBFBF}\textbf{xwinograd$\uparrow$}                                 & \cellcolor[HTML]{BFBFBF}          & \cellcolor[HTML]{BFBFBF}          & \cellcolor[HTML]{BFBFBF}\textit{\textbf{}}                    & \cellcolor[HTML]{BFBFBF}          & \cellcolor[HTML]{BFBFBF}\textit{\textbf{}}                        & \cellcolor[HTML]{BFBFBF}          & \cellcolor[HTML]{BFBFBF}\textit{\textbf{}}                   \\
\textbf{en}                                                                 & 82.15\%                           & 79.40\%                           & 78.88\%                                                       & 80.09\%                           & 79.74\%                                                           & 79.35\%                           & \textit{\textbf{79.57\%}}                                    \\
\textbf{fr}                                                                 & 71.08\%                           & 71.08\%                           & 67.47\%                                                       & 72.29\%                           & \textit{\textbf{73.49\%}}                                         & 65.06\%                           & \textit{\textbf{67.47\%}}                                    \\
\textbf{pt}                                                                 & 76.81\%                           & 74.14\%                           & \textit{\textbf{75.29\%}}                                     & 71.48\%                           & \textit{\textbf{74.14\%}}                                         & 69.20\%                           & \textit{\textbf{72.24\%}}                                    \\
\textbf{zh}                                                                 & 74.40\%                           & 74.40\%                           & \textit{\textbf{75.79\%}}                                     & 74.40\%                           & \textit{\textbf{75.20\%}}                                         & 71.23\%                           & \textit{\textbf{73.81\%}}                                    \\
\textbf{Average}                                                            & 76.11\%                           & 74.76\%                           & 74.36\%                                                       & 74.57\%                           & \textit{\textbf{75.64\%}}                                         & 71.21\%                           & \textit{\textbf{73.27\%}}                                    \\
\cellcolor[HTML]{BFBFBF}\textbf{pawsx$\uparrow$}                                     & \cellcolor[HTML]{BFBFBF}          & \cellcolor[HTML]{BFBFBF}          & \cellcolor[HTML]{BFBFBF}\textit{\textbf{}}                    & \cellcolor[HTML]{BFBFBF}          & \cellcolor[HTML]{BFBFBF}\textit{\textbf{}}                        & \cellcolor[HTML]{BFBFBF}          & \cellcolor[HTML]{BFBFBF}\textit{\textbf{}}                   \\
\textbf{en}                                                                 & 61.30\%                           & 53.60\%                           & \textit{\textbf{54.75\%}}                                     & 57.50\%                           & \textit{\textbf{58.25\%}}                                         & 56.75\%                           & \textit{\textbf{58.60\%}}                                    \\
\textbf{es}                                                                 & 59.35\%                           & 51.75\%                           & \textit{\textbf{54.05\%}}                                     & 54.10\%                           & \textit{\textbf{56.60\%}}                                         & 57.95\%                           & 56.10\%                                                      \\
\textbf{fr}                                                                 & 50.90\%                           & 47.45\%                           & 46.45\%                                                       & 50.85\%                           & 47.10\%                                                           & 52.30\%                           & 48.60\%                                                      \\
\textbf{zh}                                                                 & 47.35\%                           & 45.05\%                           & \textit{\textbf{45.45\%}}                                     & 45.70\%                           & \textit{\textbf{47.45\%}}                                         & 49.10\%                           & \textit{\textbf{50.00\%}}                                    \\
\textbf{Average}                                                            & 54.73\%                           & 49.46\%                           & \textit{\textbf{50.18\%}}                                     & 52.04\%                           & \textit{\textbf{52.35\%}}                                         & 54.03\%                           & 53.33\%                                                      \\
\cellcolor[HTML]{BFBFBF}\textbf{xnli$\uparrow$}                                      & \cellcolor[HTML]{BFBFBF}          & \cellcolor[HTML]{BFBFBF}          & \cellcolor[HTML]{BFBFBF}\textit{\textbf{}}                    & \cellcolor[HTML]{BFBFBF}          & \cellcolor[HTML]{BFBFBF}\textit{\textbf{}}                        & \cellcolor[HTML]{BFBFBF}          & \cellcolor[HTML]{BFBFBF}\textit{\textbf{}}                   \\
\textbf{ar}                                                                 & 33.83\%                           & 33.67\%                           & \textit{\textbf{33.91\%}}                                     & 34.89\%                           & 34.51\%                                                           & 33.67\%                           & \textit{\textbf{34.75\%}}                                    \\
\textbf{en}                                                                 & 53.91\%                           & 52.20\%                           & \textit{\textbf{52.59\%}}                                     & 53.49\%                           & \textit{\textbf{53.49\%}}                                         & 52.73\%                           & \textit{\textbf{52.93\%}}                                    \\
\textbf{es}                                                                 & 48.70\%                           & 48.14\%                           & 47.47\%                                                       & 45.13\%                           & \textit{\textbf{46.81\%}}                                         & 46.63\%                           & \textit{\textbf{47.54\%}}                                    \\
\textbf{fr}                                                                 & 49.68\%                           & 43.57\%                           & \textit{\textbf{48.38\%}}                                     & 46.29\%                           & \textit{\textbf{49.00\%}}                                         & 48.58\%                           & \textit{\textbf{48.62\%}}                                    \\
\textbf{hi}                                                                 & 46.51\%                           & 42.63\%                           & \textit{\textbf{44.51\%}}                                     & 40.60\%                           & \textit{\textbf{45.97\%}}                                         & 44.19\%                           & \textit{\textbf{46.63\%}}                                    \\
\textbf{sw}                                                                 & 37.92\%                           & 38.36\%                           & 37.80\%                                                       & 37.35\%                           & 36.29\%                                                           & 36.63\%                           & \textit{\textbf{37.33\%}}                                    \\
\textbf{ur}                                                                 & 42.10\%                           & 39.82\%                           & \textit{\textbf{40.54\%}}                                     & 40.42\%                           & 39.58\%                                                           & 38.42\%                           & \textit{\textbf{41.98\%}}                                    \\
\textbf{vi}                                                                 & 47.05\%                           & 45.99\%                           & \textit{\textbf{46.35\%}}                                     & 42.46\%                           & \textit{\textbf{44.89\%}}                                         & 44.29\%                           & \textit{\textbf{46.09\%}}                                    \\
\textbf{zh}                                                                 & 35.43\%                           & 35.31\%                           & 33.99\%                                                       & 34.57\%                           & 34.21\%                                                           & 35.27\%                           & 34.71\%                                                      \\
\textbf{Average}                                                            & 43.90\%                           & 42.19\%                           & \textit{\textbf{42.84\%}}                                     & 41.69\%                           & \textit{\textbf{42.75\%}}                                         & 42.27\%                           & 41.35\%                                                      \\
\cellcolor[HTML]{BFBFBF}\textbf{Average$\uparrow$}                                    & \cellcolor[HTML]{BFBFBF}57.63\%   & \cellcolor[HTML]{BFBFBF}55.36\%   & \cellcolor[HTML]{BFBFBF}\textit{\textbf{55.49\%}}             & \cellcolor[HTML]{BFBFBF}55.38\%   & \cellcolor[HTML]{BFBFBF}\textit{\textbf{56.13\%}}                 & \cellcolor[HTML]{BFBFBF}55.59\%   & \cellcolor[HTML]{BFBFBF}\textit{\textbf{57.08\%}}           
\end{tabular}
\caption{0-shot task performance of BLOOM-7b1 with different model compression methods. }
\label{0-shot7b1}
\end{table*}

\begin{table}[]
\centering
\small
\begin{tabular}{lll}
\hline
\textbf{\begin{tabular}[c]{@{}l@{}}Compression \\ Methods\end{tabular}} & \textbf{\begin{tabular}[c]{@{}l@{}}Average\\ 0-shot Task\\ Accuracy$\uparrow$\end{tabular}} & \textbf{\begin{tabular}[c]{@{}l@{}}Average \\ ppl$\downarrow$\end{tabular} }   \\  \hline
\textbf{Wanda}                                                          & 55.36\%                                                                            & 64.70                   \\
\rowcolor[HTML]{D9D9D9} 
\textbf{\begin{tabular}[c]{@{}l@{}}Wanda+\\ Equal MBS\end{tabular}}     & 55.20\%                                                                            & \textit{24.97}          \\
\rowcolor[HTML]{BFBFBF} 
\textbf{\begin{tabular}[c]{@{}l@{}}Wanda+\\ MBS\end{tabular}}           & \textit{\textbf{55.49\%}}                                                          & \textit{\textbf{26.28}} \\
\textbf{SparseGPT}                                                      & 55.38\%                                                                            & 59.84                   \\
\rowcolor[HTML]{D9D9D9} 
\textbf{\begin{tabular}[c]{@{}l@{}}SparseGPT+\\ Equal MBS\end{tabular}} & \textit{55.86\%}                                                                   & \textit{22.62}          \\
\rowcolor[HTML]{BFBFBF} 
\textbf{\begin{tabular}[c]{@{}l@{}}SparseGPT+\\ MBS\end{tabular}}       & \textit{\textbf{56.13\%}}                                                          & \textit{\textbf{23.82}} \\
\textbf{GPTQ}                                                           & 55.59\%                                                                            & 31.17                   \\
\rowcolor[HTML]{D9D9D9} 
\textbf{\begin{tabular}[c]{@{}l@{}}GPTQ+\\ Equal MBS\end{tabular}}      & \textit{56.52\%}                                                                   & \textit{23.15}          \\
\rowcolor[HTML]{BFBFBF} 
\textbf{\begin{tabular}[c]{@{}l@{}}GPTQ+\\ MBS\end{tabular}}            & \textit{\textbf{57.08\%}}                                                          & \textit{\textbf{24.26}} \\ \hline
\end{tabular}
\caption{Performance of \textit{Equal MBS}, where an equal number of segments are sampled from each language. }
\label{equalMBS}
\end{table}

We conducted our MBS sampling technique to compress both BLOOM-7b1 and BLOOM-560m models, using GPTQ, SparseGPT, and Wanda. The trends observed in the results for these two models are similar. For the sake of better formatting, we will present the results for the 7b1 model in the main text and provide the results for the 560M model in the appendices.

\textbf{Perplexity.} Figure \ref{ppl_mono_7b1} presents the evaluation of perplexity for each language after compression on the BLOOM-7b1 model. The baselines consist of monolingual compression using English-only calibration data. 

\begin{enumerate}
\setlength{\itemsep}{0pt}
\setlength{\parsep}{0pt}
\setlength{\parskip}{0pt}
    \item Across \textbf{various compression methods}, the MBS sampling technique consistently leads to minimal increases in perplexity. This holds true whether we utilize Wanda or SparseGPT for pruning or GPTQ for quantization. 
    \item For \textbf{underrepresented languages} (located on the right side of the axis), MBS can notably reduce the increase in perplexity after compression, thus preserving the model's capacity for lower-resourced languages, \textbf{leaving no languages behind}.
    \item Even for \textbf{the most well-represented language}, specifically English (on both datasets "en" and "wikitext2"), using MBS sampling introduces a \textbf{lower} perplexity than its monolingual English-centric sampling counterpart. 
\end{enumerate}

\textbf{Zero shot tasks.} Table \ref{0-shot7b1} provides an overview of the performance of zero-shot tasks after the compression process. The results demonstrate that, in the majority of tasks, utilizing MBS sampling yields superior performance compared to other sampling techniques. Furthermore, the performance after compression closely approximates that of the dense model, highlighting the effectiveness of our approach.

\textbf{Equal MBS.} Table \ref{equalMBS} provides the results for \textit{Equal MBS}, where an equal number of samples are taken from each language. While Equal MBS is not the optimal setting, it generally improves the performance of the compressed model. This demonstrates that even without access to the distribution of languages in the training set, Equal MBS can still enhance compression results for the chosen languages, showcasing the versatility of our method.

\subsection{Monolingual Compression Study}
\label{sec:mmc}
\subsubsection{Factor 1: Proportion in training data}

\begin{figure*}[htb!]
\centering
\includegraphics[width=\textwidth]{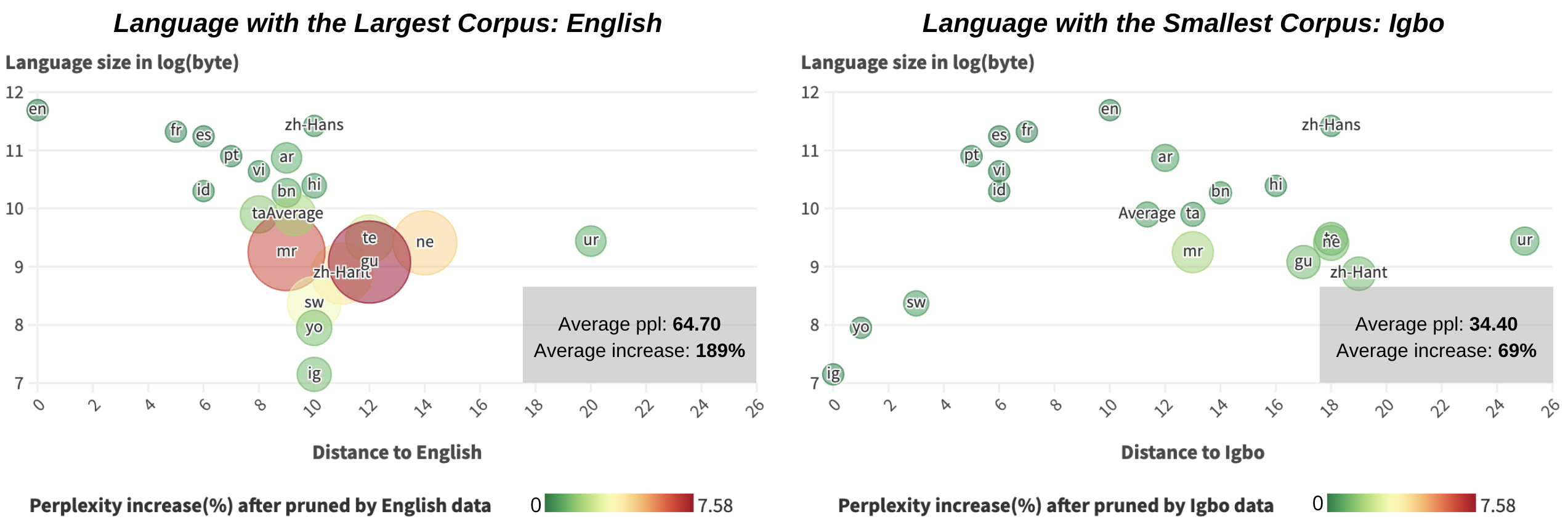}

\caption{Monolingual pruning results using Wanda with calibration data in English or Igbo. The size of each bubble corresponds to the magnitude of the increase in perplexity for the model in that particular language, while the vertical axis represents the size of training data in log(bytes) from the language in the training set of BLOOM. \textbf{The languages with a smaller proportion in the training set experience a greater increase in perplexity.}}
\label{corpus}
\end{figure*}

To investigate how the proportion of a language in the training data affects compression results, we selected English (\texttt{en}) and Igbo (\texttt{ig}), which have the largest and smallest proportions in the training data among the languages in our experiments, respectively. The results are presented in Figure \ref{corpus}.

It is evident that if we use only English as our calibration data, it significantly impacts less well-represented languages, causing substantial increases in perplexity, particularly for Marathi (\texttt{mr}) and Gujarati (\texttt{gu}). However, for better-represented languages, English (\texttt{en}) has a relatively smaller influence, as observed with Chinese simplified (\texttt{zh-Hans}), French (\texttt{fr}), and Spanish (\texttt{es}). Conversely, when we use only Igbo as our calibration data, the increase in perplexity for the other languages is relatively small. \textbf{Clearly, languages with a lower representation in the training set tend to experience a more substantial increase in perplexity. }

\subsubsection{Factor 2: Similarity between languages}

We calculated the cosine similarity of $||\mathbf{X}_n||_2^2$ for different languages using BLOOM-7b1, and then converted this similarity into degrees. This allowed us to create a distance map between languages. To visualize the relative positions of different languages, we employed Multidimensional scaling \citep{mead1992review} and generated a 2-dimensional figure (Figure \ref{cos_sim}). The original distance map is included in the appendices.

Upon observing this graph, we can identify some interesting clusters. Typically, languages from different language families tend to form distinct clusters. For instance, there is a cluster comprising Indo-European languages such as English, Spanish, French, and Portuguese, a cluster for Chinese simplified and Chinese traditional, both of which are Chinese languages, and another cluster consisting of Niger-Congo languages like Yoruba, Igbo, and Swahili. This clustering may be attributed to the following factors:
\begin{itemize}
\setlength{\itemsep}{0pt}
\setlength{\parsep}{0pt}
\setlength{\parskip}{0pt}
    \item \textbf{Shared Grammar Structure}: Languages within the same language family often share similar grammar structures.
    \item \textbf{Shared Tokens}: During the tokenization process, these languages frequently share tokens, including prefixes, suffixes, and other word-building elements.
\end{itemize}

\begin{figure}[htb!]
\centering    \includegraphics[width=0.45\textwidth]{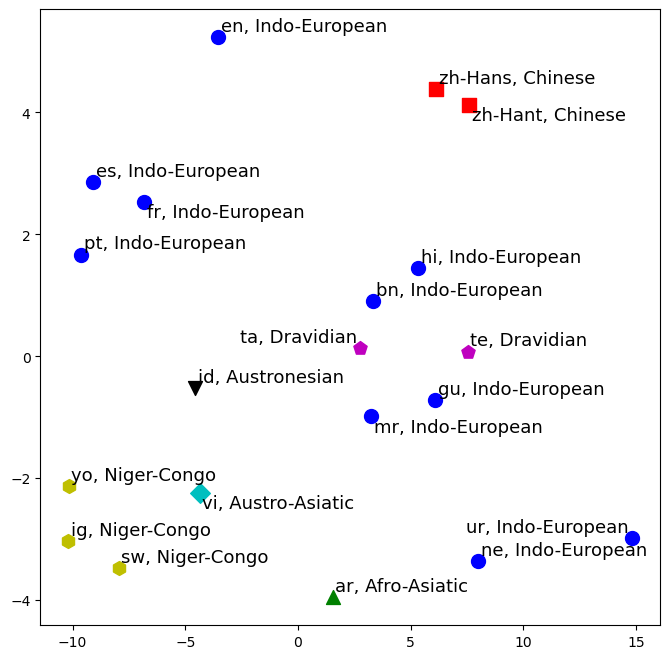}
    \caption{Distance map of different languages associated with their corresponding language families.
    We can see that languages with the same family cluster together from this map.}
    \label{cos_sim}
\end{figure}

\begin{figure*}[htb!]
\centering
\includegraphics[width=\textwidth]{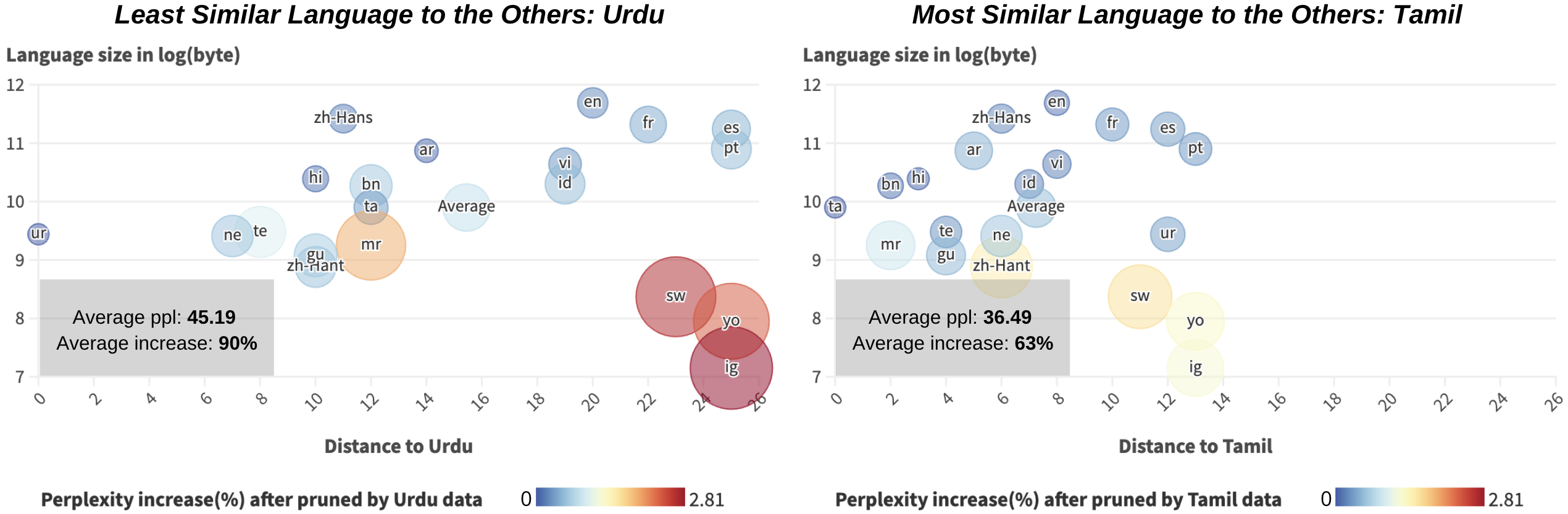}

\caption{Similarly to Figure \ref{corpus}, but focusing on Urdu or Tamil. \textbf{The languages less similar to the calibration language experience a greater increase in perplexity.}}
\label{similarity}
\end{figure*}

To investigate how language similarity impacts compression outcomes, we chose to examine two extreme cases: Tamil (\texttt{ta}), which is the language that is "closest" to all other languages (with an average distance of 7.25), and Urdu (\texttt{ur}), which is the language that is "farthest" from all other languages (with an average distance of 15.45). The results for Wanda are displayed in Figure \ref{similarity}, while the results for SparseGPT, which exhibit similar patterns to those of Wanda, are provided in the appendices.

It is evident that when we use Tamil (\texttt{ta}) as our sole calibration data, the increase in perplexity for other languages is relatively small, especially for languages that are "closer" to Tamil, such as Bengali (\texttt{bn}) and Hindi (\texttt{hi}). Conversely, when Urdu (\texttt{ur}) serves as our sole calibration data, the increase in perplexity for other languages is relatively significant on average. The consistent pattern across all four graphs reveals that \textbf{languages more distant from the language being compressed tend to exhibit a more significant increase in perplexity.}


An intriguing case study can be conducted on Chinese simplified and Chinese traditional. Despite their close proximity on the language map, they significantly differ in corpus size. Chinese simplified enjoys a much larger proportion, resulting in a more pronounced impact on Chinese traditional after the compression process, while Chinese simplified remains relatively unaffected. These experiments demonstrate the validity and accuracy of our theory.

\section{Related Work}

\textbf{Large Language Model.} Large Language Models\cite{zhao2023survey} like GPT-4\cite{openai2024gpt4}, LLaMA\cite{touvron2023llama} and OPT\cite{zhang2022opt}, which have revolutionized Natural Language Processing through their ability to understand and generate nuanced text. Alongside, multilingual language models\cite{doddapaneni2021primer} such as BLOOM\cite{bigscience_workshop_2022} and XLM-R\cite{conneau2020unsupervised} are breaking language barriers by learning universal representations from texts across numerous languages. These developments underscore a significant shift towards creating more versatile and inclusive NLP systems, with research focusing on architectural innovations, training efficiencies, and cross-lingual capabilities to enhance global digital interaction.

We would like to emphasize that \textit{MBS can be applied to any model compression method that utilizes calibration data, particularly methods based on the OBS/OBD framework}, where the approximation of second-derivative information is required. Thanks to a survey on model compression for large language models by \citep{zhu2023survey}, we examined the state-of-the-art model compression methods for large language models, and we found that our MBS is useful for almost all of them.

Pruning and quantization are two major model compression methods for LLMs. 

\textbf{Pruning.} Pruning reduces model size and complexity by eliminating unnecessary or redundant components. It can be categorized into \textbf{structured pruning}, where higher-granularity structures like rows or columns of weight matrices are removed, and \textbf{unstructured pruning}, which eliminates individual weights, leading to irregular sparse structures. In the domain of \textbf{unstructured pruning}, MBS can be applied to Wanda \citep{sun2023simple} and SparseGPT \citep{frantar2023sparsegpt} that we presented in the background section, and also LoRAPrune \citep{zhang2023loraprune}. In the \textbf{structured pruning} domain, MBS can empower LLM-Pruner \citep{ma2023llmpruner}.

\textbf{Quantization.} Quantization involves converting floating-point numbers into lower bit-level representations, integers, or other discrete forms and can be categorized into \textbf{Quantization-Aware Training} and \textbf{Post-Training Quantization}. MBS finds numerous applications in quantization, particularly in \textbf{post-training quantization}. In post-training quantization, certain approaches focus on quantizing \textbf{only the
weights} of LLMs. Among these methods, MBS can be applied to AWQ \citep{lin2023awq}, GPTQ \citep{frantar2023gptq}, OWQ \citep{lee2023owq}, SpQR \citep{dettmers2023spqr}, SqueezeLLM \citep{kim2023squeezellm}, QuIP \citep{chee2023quip}, and SignRound \citep{cheng2023optimize}. Some other methods try to quantize \textbf{both weights
and activations} of LLMs. Among them, MBS can be applied to SmoothQuant \citep{xiao2023smoothquant}, RPTQ \citep{yuan2023rptq}, OliVe \citep{Guo_2023}, ZeroQuant-V2 \citep{yao2023zeroquantv2}, Outlier Suppression+ \citep{wei2023outlier}, FPTQ \citep{li2023fptq}, QuantEase \citep{behdin2023quantease}, and OmniQuant \citep{shao2023omniquant}.

\section{Conclusions}

In summary, the Multilingual Brain Surgeon (MBS) is a groundbreaking approach for improving multilingual LLMs. It tackles the English-centric bias in existing techniques and enhances LLM performance after compressing. Our experiments on the BLOOM model highlight the effectiveness of MBS, benefiting pruning and quantization methods like SparseGPT, Wanda, and GPTQ.

We also studied language interaction during compression, finding that language proportion in the training dataset and language similarity are crucial factors. Languages with larger proportion are less affected by compression, while similar languages perform better when only one language is used in calibration data. Our proposed similarity measure accurately predicts performance drops in such scenarios.

This research not only enhances the practicality of multilingual LLMs compression methods but also maintains language coverage, making multilingual NLP applications more inclusive and powerful.

\section*{Acknowledgments}
This work is funded by the China NSFC Projects (92370206, U23B2057, 62106142 and 62120106006) and Shanghai Municipal Science and Technology Major Project (2021SHZDZX0102).

\nocite{*}
\bibliography{lrec-coling2024-example}

\newpage
\appendix

\section{Details of Calibration Data}

\begin{table}[h]
\begin{tabular}{llll}
\hline
\rowcolor{gray!40}
Language & \begin{tabular}[c]{@{}l@{}}Size in Bytes\\ in BLOOM\\ training data\end{tabular} & \begin{tabular}[c]{@{}l@{}}MBS \\ sampling\end{tabular} & \begin{tabular}[c]{@{}l@{}}Equal \\ sampling\end{tabular} \\ \hline
en       & 4.85E+11                                                                         & 87           & 13             \\
zh-Hans  & 2.61E+11                                                                         & 47           & 13             \\
fr       & 2.08E+11                                                                         & 37           & 13             \\
es       & 1.75E+11                                                                         & 31           & 13             \\
pt       & 7.93E+10                                                                         & 14           & 13             \\
ar       & 7.49E+10                                                                         & 13           & 13             \\
vi       & 4.37E+10                                                                         & 7            & 13             \\
hi       & 2.46E+10                                                                         & 4            & 13             \\
id       & 2.00E+10                                                                         & 3            & 13             \\
bn       & 1.86E+10                                                                         & 3            & 13             \\
ta       & 7.99E+09                                                                         & 1            & 13             \\
te       & 2.99E+09                                                                         & 1            & 13             \\
ur       & 2.78E+09                                                                         & 1            & 13             \\
ne       & 2.55E+09                                                                         & 1            & 13             \\
mr       & 1.78E+09                                                                         & 1            & 13             \\
gu       & 1.20E+09                                                                         & 1            & 13             \\
zh-Hant  & 7.62E+08                                                                         & 1            & 12             \\
sw       & 2.36E+08                                                                         & 1            & 12             \\
yo       & 8.97E+07                                                                         & 1            & 12             \\
ig       & 1.41E+07                                                                         & 1            & 12         \\ \hline   
\end{tabular}
\caption{The number of segments taken from each language by each sampling method. }
\label{pruningdetails}
\end{table}

The number of segments taken from each language is detailed in Table \ref{pruningdetails}. We rounded up the segment counts for languages with fewer than one segment to ensure their representation in the calibration data. In the equal sampling scenario, to maintain comparability, some languages have one segment less than others to achieve a total of 256 segments.

\section{MBS results tables}

\begin{table*}[htb!] 
\centering
\begin{tabular}{llllllllll}
\hline
\textbf{Dataset} & \textbf{Dense} & \textbf{Wanda} & \textbf{$\uparrow$}                     & \textbf{SparseGPT} & \textbf{$\uparrow$}                     & \textbf{\begin{tabular}[c]{@{}l@{}}MBS + \\ Wanda\end{tabular}} &             \textbf{$\uparrow$}                 & \textbf{\begin{tabular}[c]{@{}l@{}}MBS + \\ SparseGPT\end{tabular}} &          \textbf{$\uparrow$}                    \\
\hline
en               & 13.68          & 15.67          & \cellcolor[HTML]{88C87D}15\%  & 14.92              & \cellcolor[HTML]{67BF7B}9\%   & 15.55                & \cellcolor[HTML]{83C77C}14\% & 15.01           & \cellcolor[HTML]{6BC07B}10\% \\
zh-Hans          & 23.70          & 34.75          & \cellcolor[HTML]{FFE984}47\%  & 35.56              & \cellcolor[HTML]{FFE984}50\%  & 26.59                & \cellcolor[HTML]{7AC47C}12\% & 25.87           & \cellcolor[HTML]{68BF7B}9\%  \\
fr               & 9.59           & 13.16          & \cellcolor[HTML]{FFEB84}37\%  & 13.41              & \cellcolor[HTML]{FFEB84}40\%  & 10.68                & \cellcolor[HTML]{75C37C}11\% & 10.39           & \cellcolor[HTML]{63BE7B}8\%  \\
es               & 10.71          & 13.82          & \cellcolor[HTML]{DFE182}29\%  & 13.75              & \cellcolor[HTML]{DBE081}28\%  & 11.91                & \cellcolor[HTML]{74C37C}11\% & 11.59           & \cellcolor[HTML]{63BE7B}8\%  \\
pt               & 10.97          & 14.58          & \cellcolor[HTML]{F6E883}33\%  & 17.37              & \cellcolor[HTML]{FFE784}58\%  & 12.25                & \cellcolor[HTML]{77C37C}12\% & 11.96           & \cellcolor[HTML]{67BF7B}9\%  \\
ar               & 14.40          & 29.19          & \cellcolor[HTML]{FFDF82}103\% & 25.33              & \cellcolor[HTML]{FFE483}76\%  & 16.45                & \cellcolor[HTML]{86C87D}14\% & 15.88           & \cellcolor[HTML]{6FC17B}10\% \\
vi               & 10.16          & 14.76          & \cellcolor[HTML]{FFEA84}45\%  & 15.00              & \cellcolor[HTML]{FFE984}48\%  & 11.59                & \cellcolor[HTML]{86C87D}14\% & 11.24           & \cellcolor[HTML]{70C27B}11\% \\
hi               & 10.96          & 18.26          & \cellcolor[HTML]{FFE683}67\%  & 19.09              & \cellcolor[HTML]{FFE483}74\%  & 12.52                & \cellcolor[HTML]{86C87D}14\% & 12.19           & \cellcolor[HTML]{74C37C}11\% \\
id               & 20.48          & 29.37          & \cellcolor[HTML]{FFEA84}43\%  & 37.27              & \cellcolor[HTML]{FFE383}82\%  & 23.76                & \cellcolor[HTML]{91CB7D}16\% & 22.97           & \cellcolor[HTML]{7AC47C}12\% \\
bn               & 17.27          & 33.37          & \cellcolor[HTML]{FFE182}93\%  & 40.50              & \cellcolor[HTML]{FFDA81}134\% & 20.29                & \cellcolor[HTML]{99CD7E}17\% & 19.51           & \cellcolor[HTML]{7EC67C}13\% \\
ta               & 16.55          & 42.23          & \cellcolor[HTML]{FED680}155\% & 44.10              & \cellcolor[HTML]{FED480}167\% & 20.10                & \cellcolor[HTML]{B2D47F}21\% & 19.34           & \cellcolor[HTML]{96CC7D}17\% \\
te               & 18.10          & 64.97          & \cellcolor[HTML]{FDC37D}259\% & 69.20              & \cellcolor[HTML]{FDBF7C}282\% & 24.59                & \cellcolor[HTML]{FFEB84}36\% & 22.05           & \cellcolor[HTML]{B3D57F}22\% \\
ur               & 13.26          & 27.03          & \cellcolor[HTML]{FFDF82}104\% & 30.56              & \cellcolor[HTML]{FFDA81}130\% & 15.83                & \cellcolor[HTML]{A5D17E}19\% & 15.10           & \cellcolor[HTML]{84C77C}14\% \\
ne               & 27.22          & 152.67         & \cellcolor[HTML]{FB9F76}461\% & 148.00             & \cellcolor[HTML]{FCA276}444\% & 34.90                & \cellcolor[HTML]{DAE081}28\% & 32.91           & \cellcolor[HTML]{AED37F}21\% \\
mr               & 23.07          & 176.78         & \cellcolor[HTML]{F97A6F}666\% & 144.65             & \cellcolor[HTML]{FB9373}527\% & 32.25                & \cellcolor[HTML]{FFEB84}40\% & 28.91           & \cellcolor[HTML]{C8DB80}25\% \\
gu               & 21.52          & 184.62         & \cellcolor[HTML]{F8696B}758\% & 118.48             & \cellcolor[HTML]{FBA176}450\% & 30.84                & \cellcolor[HTML]{FFEA84}43\% & 26.97           & \cellcolor[HTML]{C9DB80}25\% \\
zh-Hant          & 21.84          & 113.34         & \cellcolor[HTML]{FCA677}419\% & 102.26             & \cellcolor[HTML]{FCB079}368\% & 24.96                & \cellcolor[HTML]{86C87D}14\% & 24.30           & \cellcolor[HTML]{74C37C}11\% \\
sw               & 34.35          & 145.54         & \cellcolor[HTML]{FDB87B}324\% & 135.84             & \cellcolor[HTML]{FDBD7B}295\% & 54.32                & \cellcolor[HTML]{FFE784}58\% & 44.23           & \cellcolor[HTML]{DDE182}29\% \\
yo               & 53.29          & 128.12         & \cellcolor[HTML]{FED881}140\% & 126.54             & \cellcolor[HTML]{FFD981}137\% & 79.62                & \cellcolor[HTML]{FFE984}49\% & 67.52           & \cellcolor[HTML]{D1DD81}27\% \\
ig               & 39.16          & 90.41          & \cellcolor[HTML]{FFDA81}131\% & 90.89              & \cellcolor[HTML]{FFDA81}132\% & 59.00                & \cellcolor[HTML]{FFE984}51\% & 49.10           & \cellcolor[HTML]{C9DB80}25\% \\ \hline
wikitext2        & 11.37          & 16.15          & \cellcolor[HTML]{FFEA84}42\%  & 13.91              & \cellcolor[HTML]{B7D67F}22\%  & 13.82                & \cellcolor[HTML]{B2D47F}22\% & 13.26           & \cellcolor[HTML]{95CC7D}17\% \\
                 \hline
\textbf{Average}          & 20.08          & 64.70          & 222\%                         & 59.84              & 198\%                         & \textbf{26.28}                & \textbf{31\%}                         & \textbf{23.82}           & \textbf{19\% }    \\
\hline
\end{tabular}
\caption{Perplexity for each language and their respective increases when compared to the dense BLOOM-7b1 model after pruning. Evaluation performed on XL-Sum and WikiText2 datasets. From top to bottom, languages are ranked in order from the most well-represented to the least represented.}
\end{table*}

\begin{table*}[]
\centering
\begin{tabular}{|l|l|l|l|}
\hline
\textbf{Dataset}   & \textbf{Dense} & \textbf{GPTQ} & \textbf{GPTQ+MBS} \\ \hline
\textbf{en}        & 13.68          & 15.3          & 15.37             \\ \hline
\textbf{zh-Hans}   & 23.7           & 28.69         & 26.28             \\ \hline
\textbf{fr}        & 9.59           & 10.94         & 10.46             \\ \hline
\textbf{es}        & 10.71          & 12.49         & 11.9              \\ \hline
\textbf{pt}        & 10.97          & 13.05         & 12.36             \\ \hline
\textbf{ar}        & 14.4           & 17.35         & 16.12             \\ \hline
\textbf{vi}        & 10.16          & 11.97         & 11.15             \\ \hline
\textbf{hi}        & 10.96          & 13.72         & 12.27             \\ \hline
\textbf{id}        & 20.48          & 25.36         & 23.45             \\ \hline
\textbf{bn}        & 17.27          & 22.88         & 19.83             \\ \hline
\textbf{ta}        & 16.55          & 24.5          & 19.53             \\ \hline
\textbf{te}        & 18.1           & 32.83         & 22.67             \\ \hline
\textbf{ur}        & 13.26          & 18.29         & 15.4              \\ \hline
\textbf{ne}        & 27.22          & 45.5          & 33.97             \\ \hline
\textbf{mr}        & 23.07          & 43.32         & 29.57             \\ \hline
\textbf{gu}        & 21.52          & 40.4          & 27.83             \\ \hline
\textbf{zh-Hant}   & 21.84          & 26.85         & 24.75             \\ \hline
\textbf{sw}        & 34.35          & 68.42         & 46.19             \\ \hline
\textbf{yo}        & 53.29          & 98.46         & 67.82             \\ \hline
\textbf{ig}        & 39.16          & 71.56         & 49.96             \\ \hline
\textbf{wikitext2} & 11.37          & 12.6          & 12.56             \\ \hline
\textbf{Average}   & 20.08          & 31.17         & 24.26             \\ \hline
\end{tabular}
\caption{Perplexity for each language of BLOOM-7b1 model before and after quantization. }
\end{table*}

\begin{table*}[htb]
\centering
\begin{tabular}{llllll}
\hline
Dataset   & Dense & Wanda                            & SparseGPT                        & \begin{tabular}[c]{@{}l@{}}MBS + \\ Wanda\end{tabular} & \begin{tabular}[c]{@{}l@{}}MBS + \\ SparseGPT\end{tabular} \\ \hline
en        & 25.9744  & \cellcolor[HTML]{79C47C}34.30    & \cellcolor[HTML]{63BE7B}32.02    & \cellcolor[HTML]{97CD7E}35.08  & \cellcolor[HTML]{92CB7D}32.87  \\ 
zh-Hans   & 46.1386  & \cellcolor[HTML]{C4DA80}72.92    & \cellcolor[HTML]{65BE7B}101.79   & \cellcolor[HTML]{FFEB84}62.39  & \cellcolor[HTML]{F7E883}59.28  \\ 
zh-Hant   & 43.7887  & \cellcolor[HTML]{FFE784}137.00   & \cellcolor[HTML]{69BF7B}266.49   & \cellcolor[HTML]{FFEB84}61.30  & \cellcolor[HTML]{F4E783}58.56  \\ 
fr        & 16.3204  & \cellcolor[HTML]{63BE7B}22.89    & \cellcolor[HTML]{63BE7B}24.99    & \cellcolor[HTML]{63BE7B}21.66  & \cellcolor[HTML]{63BE7B}20.37  \\ 
es        & 18.1103  & \cellcolor[HTML]{67BF7B}25.41    & \cellcolor[HTML]{63BE7B}27.94    & \cellcolor[HTML]{6CC07B}23.98  & \cellcolor[HTML]{6BC07B}22.56  \\ 
pt        & 19.1249  & \cellcolor[HTML]{6BC07B}27.47    & \cellcolor[HTML]{63BE7B}34.99    & \cellcolor[HTML]{73C27B}25.82  & \cellcolor[HTML]{71C27B}24.27  \\ 
ar        & 28.1339  & \cellcolor[HTML]{D2DE81}80.09    & \cellcolor[HTML]{FFEB84}1.13E+14 & \cellcolor[HTML]{B0D47F}41.36  & \cellcolor[HTML]{CADB80}47.57  \\ 
vi        & 20.7135  & \cellcolor[HTML]{8AC97D}43.26    & \cellcolor[HTML]{65BE7B}127.59   & \cellcolor[HTML]{86C87D}30.75  & \cellcolor[HTML]{86C87D}29.64  \\ 
hi        & 20.8597  & \cellcolor[HTML]{8CC97D}44.10    & \cellcolor[HTML]{FFEB84}1.20E+15 & \cellcolor[HTML]{80C67C}29.17  & \cellcolor[HTML]{BCD780}43.84  \\ 
id        & 42.0467  & \cellcolor[HTML]{FFEB84}109.13   & \cellcolor[HTML]{FFEB84}76038.83 & \cellcolor[HTML]{FFEB84}64.40  & \cellcolor[HTML]{FFE784}79.85  \\ 
bn        & 37.0531  & \cellcolor[HTML]{FFEB84}102.78   & \cellcolor[HTML]{F8696B}4.42E+23 & \cellcolor[HTML]{ECE582}56.65  & \cellcolor[HTML]{FFDE82}121.11 \\ 
ta        & 36.9749  & \cellcolor[HTML]{FFE183}176.25   & \cellcolor[HTML]{FFEB84}1.71E+07 & \cellcolor[HTML]{FFEB84}64.40  & \cellcolor[HTML]{FEC97E}209.90 \\ 
te        & 54.3996  & \cellcolor[HTML]{FEC87E}355.98   & \cellcolor[HTML]{FFEB84}6.96E+05 & \cellcolor[HTML]{FFDD82}116.20 & \cellcolor[HTML]{FCA777}355.54 \\ 
ur        & 29.419   & \cellcolor[HTML]{F4E883}97.65    & \cellcolor[HTML]{FFEB84}3.62E+20 & \cellcolor[HTML]{BDD880}44.60  & \cellcolor[HTML]{FFEB84}61.35  \\ 
ne        & 75.7999  & \cellcolor[HTML]{FCAC78}555.21   & \cellcolor[HTML]{FFEB84}1.27E+17 & \cellcolor[HTML]{FED781}135.67 & \cellcolor[HTML]{FECB7E}200.16 \\ 
mr        & 72.4524  & \cellcolor[HTML]{FCB37A}503.51   & \cellcolor[HTML]{FFEB84}4.25E+12 & \cellcolor[HTML]{FED580}142.76 & \cellcolor[HTML]{F8696B}619.28 \\ 
gu        & 70.6476  & \cellcolor[HTML]{FECC7E}327.11   & \cellcolor[HTML]{FFEB84}2.17E+13 & \cellcolor[HTML]{FED781}136.47 & \cellcolor[HTML]{FECC7E}197.21 \\ 
sw        & 154.7551 & \cellcolor[HTML]{F9786E}919.14   & \cellcolor[HTML]{FFEB84}6011.36  & \cellcolor[HTML]{FB9B75}357.63 & \cellcolor[HTML]{FCAB78}336.80 \\ 
yo        & 267.0194 & \cellcolor[HTML]{F8696B}1.02E+03 & \cellcolor[HTML]{BFD880}3557.88  & \cellcolor[HTML]{F8696B}542.40 & \cellcolor[HTML]{FB9173}450.00 \\ 
ig        & 185.7405 & \cellcolor[HTML]{FB9374}728.01   & \cellcolor[HTML]{84C77C}1306.30  & \cellcolor[HTML]{FB9874}370.35 & \cellcolor[HTML]{FCB279}307.76 \\ 
wikitext2 & 22.4121  & \cellcolor[HTML]{72C27B}30.58    & \cellcolor[HTML]{63BE7B}29.75    & \cellcolor[HTML]{88C87D}31.09  & \cellcolor[HTML]{87C87D}29.90  \\ \hline
Average       & 61.33    & 257.99                           & 2.11E+22                         & 114.01                         & 157.51                         \\ \hline
\end{tabular}

\caption{Perplexity for each language and their respective increases when compared to the dense BLOOM-560m model after pruning. Evaluation performed on XL-Sum and WikiText2 datasets. From top to bottom, languages are ranked in order from the most well-represented to the least represented.}
\label{560mpplmono}
\end{table*}

\textbf{Perplexity.} Table \ref{560mpplmono} showcases the perplexity evaluation for each language after pruning on the BLOOM-560m model. The observed trends align closely with those observed on the BLOOM-7b1 model. 

\textbf{Zero shot tasks.} Table \ref{0shot560m} illustrates the zero-shot task results for the pruned BLOOM-560m model. It is noticeable that the average accuracy using the MBS sampling method continues to outperform the baselines, although the results appear to exhibit more variability. This variability can be attributed to the reduced capacity of smaller models to maintain their multilingual capabilities.

\textbf{The role of parameter compensation.} We have observed a notable distinction in the effects of SparseGPT and Wanda. In monolingual pruning, SparseGPT, which involves parameter updates and employs a more precise pruning metric, appears to have a more detrimental impact on less well-represented languages. However, when we apply MBS, SparseGPT continues to outperform Wanda. This phenomenon may be attributed to the fact that smaller models are more sensitive to parameter updates. A biased Hessian matrix can exacerbate the model's divergence from the correct direction through these updates. Conversely, a correctly approximated Hessian matrix can effectively guide the pruning in the correct direction.

\begin{table*}[!htb]
\begin{tabular}{llllllll}
\hline
\textbf{0-shot task}                                              & \textbf{Dense} & \textbf{Wanda}   & \textbf{SparseGPT} & \textbf{\begin{tabular}[c]{@{}l@{}}Wanda \\ equal\end{tabular}} & \textbf{\begin{tabular}[c]{@{}l@{}}SparseGPT \\ equal\end{tabular}} & \textbf{\begin{tabular}[c]{@{}l@{}}MBS+\\ Wanda\end{tabular}} & \textbf{\begin{tabular}[c]{@{}l@{}}MBS+\\ SparseGPT\end{tabular}} \\ \rowcolor{gray!40}
\textbf{xcopa}                                                    &                &                  &                    &                                                                 &                                                                     &                                                               &                                                                   \\
\textbf{id}                                                       & 59.20\%        & 56.20\%          & 51.80\%            & 58.40\%                                                         & 57.40\%                                                             & \textbf{58.60\%}                                              & 55.80\%                                                           \\
\textbf{sw}                                                       & 51.60\%        & 52.60\%          & 52.60\%            & 51.80\%                                                         & \textbf{52.80\%}                                                    & 52.20\%                                                       & 52.20\%                                                           \\
\textbf{ta}                                                       & 55.80\%        & \textbf{57.20\%} & 54.00\%            & 56.00\%                                                         & 56.00\%                                                             & 54.80\%                                                       & 56.20\%                                                           \\
\textbf{vi}                                                       & 61.00\%        & 55.40\%          & 51.60\%            & \textbf{57.60\%}                                                & 56.20\%                                                             & 56.00\%                                                       & 55.60\%                                                           \\
\textbf{zh}                                                       & 58.60\%        & 52.80\%          & 53.00\%            & 53.40\%                                                         & 53.60\%                                                             & 53.80\%                                                       & \textbf{55.00\%}                                                  \\
\textbf{Average}                                                  & 57.24\%        & 54.84\%          & 52.60\%            & \textbf{55.44\%}                                                & 55.20\%                                                             & 55.08\%                                                       & 54.96\%                                                           \\ \rowcolor{gray!40}
\textbf{xstory\_cloze}                                            &                &                  &                    &                                                                 &                                                                     &                                                               &                                                                   \\
\textbf{ar}                                                       & 52.08\%        & 48.25\%          & \textbf{49.57\%}   & 49.24\%                                                         & 48.31\%                                                             & 48.97\%                                                       & 48.84\%                                                           \\
\textbf{en}                                                       & 61.22\%        & 57.78\%          & 59.23\%            & 56.92\%                                                         & 57.97\%                                                             & 57.51\%                                                       & \textbf{59.70\%}                                                  \\
\textbf{es}                                                       & 55.86\%        & 53.67\%          & 54.40\%            & 53.81\%                                                         & 54.14\%                                                             & 54.27\%                                                       & \textbf{55.00\%}                                                  \\
\textbf{hi}                                                       & 55.00\%        & 52.68\%          & 48.31\%            & 53.21\%                                                         & \textbf{54.00\%}                                                    & 53.08\%                                                       & 52.88\%                                                           \\
\textbf{id}                                                       & 55.53\%        & 53.14\%          & 48.31\%            & 53.14\%                                                         & 52.42\%                                                             & \textbf{53.34\%}                                              & 52.22\%                                                           \\
\textbf{sw}                                                       & 49.83\%        & 49.11\%          & 48.78\%            & 49.24\%                                                         & 49.24\%                                                             & \textbf{49.44\%}                                              & 48.51\%                                                           \\
\textbf{te}                                                       & 55.72\%        & 54.33\%          & 53.28\%            & 54.80\%                                                         & \textbf{55.46\%}                                                    & 54.07\%                                                       & \textbf{55.46\%}                                                  \\
\textbf{zh}                                                       & 54.53\%        & 51.95\%          & 51.29\%            & 51.56\%                                                         & 52.68\%                                                             & 51.95\%                                                       & \textbf{53.54\%}                                                  \\
\textbf{Average}                                                  & 54.97\%        & 52.61\%          & 51.65\%            & 52.74\%                                                         & 53.03\%                                                             & 52.83\%                                                       & \textbf{53.27\%}                                                  \\ \rowcolor{gray!40}
\textbf{xwinograd}                                                &                &                  &                    &                                                                 &                                                                     &                                                               &                                                                   \\
\textbf{en}                                                       & 65.89\%        & 61.98\%          & 62.58\%            & 62.28\%                                                         & \textbf{63.44\%}                                                    & 62.02\%                                                       & 62.92\%                                                           \\
\textbf{fr}                                                       & 60.24\%        & 56.63\%          & 51.81\%            & 56.63\%                                                         & 56.63\%                                                             & \textbf{59.04\%}                                              & 57.83\%                                                           \\
\textbf{pt}                                                       & 60.08\%        & 55.51\%          & \textbf{60.46\%}   & 54.75\%                                                         & 57.41\%                                                             & 55.13\%                                                       & 59.70\%                                                           \\
\textbf{zh}                                                       & 67.66\%        & 66.27\%          & 66.87\%            & 66.87\%                                                         & 63.49\%                                                             & 66.07\%                                                       & \textbf{69.64\%}                                                  \\
\textbf{Average}                                                  & 63.47\%        & 60.10\%          & 60.43\%            & 60.13\%                                                         & 60.24\%                                                             & 60.57\%                                                       & \textbf{62.52\%}                                                  \\ \rowcolor{gray!40}
\textbf{pawsx}                                                    &                &                  &                    &                                                                 &                                                                     &                                                               &                                                                   \\
\textbf{en}                                                       & 52.00\%        & 49.90\%          & 48.60\%            & 49.15\%                                                         & 49.85\%                                                             & 47.60\%                                                       & \textbf{50.85\%}                                                  \\
\textbf{es}                                                       & 53.25\%        & 48.75\%          & 48.85\%            & \textbf{51.60\%}                                                & 48.75\%                                                             & 50.70\%                                                       & 50.80\%                                                           \\
\textbf{fr}                                                       & 47.95\%        & 47.70\%          & \textbf{48.45\%}   & 46.45\%                                                         & 46.75\%                                                             & 46.45\%                                                       & 45.35\%                                                           \\
\textbf{zh}                                                       & 45.20\%        & 45.15\%          & 44.85\%            & 45.50\%                                                         & \textbf{45.70\%}                                                    & 45.55\%                                                       & 44.75\%                                                           \\
\textbf{Average}                                                  & 49.60\%        & 47.88\%          & 47.69\%            & \textbf{48.18\%}                                                & 47.76\%                                                             & 47.58\%                                                       & 47.94\%                                                           \\ \rowcolor{gray!40}
\textbf{xnli}                                                     &                &                  &                    &                                                                 &                                                                     &                                                               &                                                                   \\
\textbf{ar}                                                       & 33.35\%        & 33.47\%          & 33.55\%            & 33.59\%                                                         & 33.57\%                                                             & 33.41\%                                                       & \textbf{33.67\%}                                                  \\
\textbf{en}                                                       & 49.50\%        & \textbf{46.53\%} & 45.89\%            & 45.43\%                                                         & 46.09\%                                                             & 45.31\%                                                       & 46.37\%                                                           \\
\textbf{es}                                                       & 45.23\%        & \textbf{45.71\%} & 41.72\%            & 43.11\%                                                         & 42.87\%                                                             & 44.45\%                                                       & 42.95\%                                                           \\
\textbf{fr}                                                       & 45.29\%        & 45.51\%          & 42.20\%            & 45.07\%                                                         & 45.27\%                                                             & \textbf{45.71\%}                                              & 43.75\%                                                           \\
\textbf{hi}                                                       & 40.84\%        & 38.64\%          & 33.35\%            & 37.90\%                                                         & 36.45\%                                                             & \textbf{38.78\%}                                              & 34.77\%                                                           \\
\textbf{sw}                                                       & 33.17\%        & \textbf{33.65\%} & 33.51\%            & 33.45\%                                                         & 33.21\%                                                             & 33.63\%                                                       & 33.29\%                                                           \\
\textbf{ur}                                                       & 37.13\%        & 33.75\%          & 33.27\%            & \textbf{35.23\%}                                                & 34.77\%                                                             & 35.11\%                                                       & 34.01\%                                                           \\
\textbf{vi}                                                       & 40.52\%        & \textbf{40.28\%} & 33.05\%            & 38.44\%                                                         & 36.61\%                                                             & 39.14\%                                                       & 37.15\%                                                           \\
\textbf{zh}                                                       & 33.95\%        & 33.33\%          & 33.45\%            & 33.29\%                                                         & 33.47\%                                                             & 33.29\%                                                       & \textbf{33.61\%}                                                  \\
\textbf{Average}                                                  & 39.89\%        & \textbf{38.99\%} & 36.67\%            & 38.39\%                                                         & 38.04\%                                                             & 38.76\%                                                       & 37.73\%                                                           \\
\rowcolor{gray!40}
\textbf{\begin{tabular}[c]{@{}l@{}}Total \\ Average\end{tabular}} & 51.24\%        & 49.26\%          & 47.95\%            & 49.26\%                                                         & 49.15\%                                                             & \textbf{49.31\%}                                                       & \textbf{49.41\%}                                                 
\end{tabular}
\caption{0-shot tasks performance on each task of BLOOM-560m pruned model.}
\label{0shot560m}
\end{table*}

\section{Monolingual pruning results}

\begin{figure*} [htb]
\centering
    \includegraphics[width=0.5\textwidth]{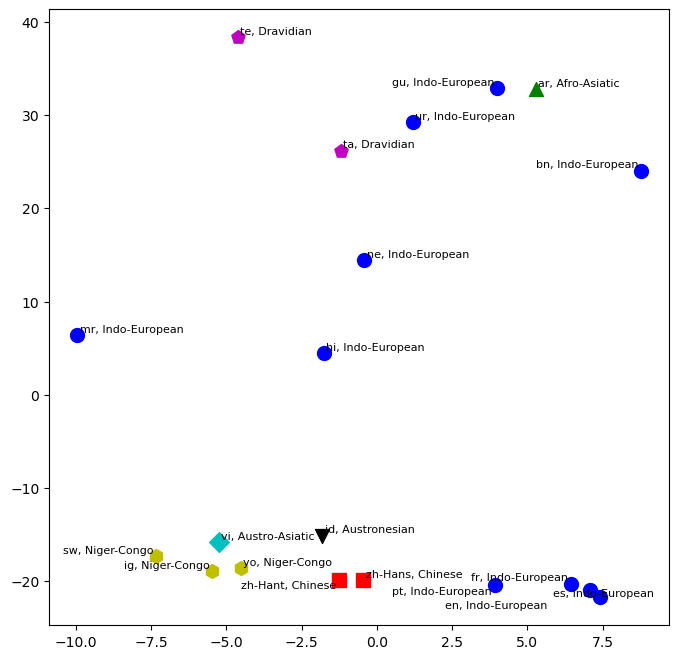}
    \caption{The graph illustrates the relative positions of different languages. Different dot shapes represent different language families. The closer they are on the graph, the more similar they are to each other.}
    \label{cos_sim_560m}
\end{figure*}

\textbf{Distance map of BLOOM-560m model.} The distance map depicting the relationships between languages in the BLOOM-560m model is depicted in Figure \ref{cos_sim_560m}. We can discern a similar clustering pattern to that observed in the BLOOM-7b1 model.

The original distance matrices are provided in the following tables(\ref{dist7b1}, \ref{dist560m}).

\begin{table*}[]
\begin{tabular}{llllllllll}
\hline
\textbf{Dataset}   & \textbf{Dense} & \textbf{en} & \textbf{$\uparrow$}                     & \textbf{ig} & \textbf{$\uparrow$}                     & \textbf{ta} & \textbf{$\uparrow$}                     & \textbf{ur} & \textbf{$\uparrow$}                     \\  \hline
\textbf{en}        & 13.68          & 15.67       & \cellcolor[HTML]{84C77C}15\%  & 17.61       & \cellcolor[HTML]{B9D67F}29\%  & 17.20       & \cellcolor[HTML]{AED37F}26\%  & 18.93       & \cellcolor[HTML]{DDE182}38\%  \\
\textbf{zh-Hans}   & 23.70          & 34.75       & \cellcolor[HTML]{FCEA83}47\%  & 33.97       & \cellcolor[HTML]{EFE683}43\%  & 31.99       & \cellcolor[HTML]{D0DD81}35\%  & 31.70       & \cellcolor[HTML]{CCDC81}34\%  \\
\textbf{fr}        & 9.59           & 13.16       & \cellcolor[HTML]{D8E081}37\%  & 12.70       & \cellcolor[HTML]{C7DA80}32\%  & 13.94       & \cellcolor[HTML]{F7E883}45\%  & 14.84       & \cellcolor[HTML]{FFEA84}55\%  \\
\textbf{es}        & 10.71          & 13.82       & \cellcolor[HTML]{BAD780}29\%  & 14.16       & \cellcolor[HTML]{C6DA80}32\%  & 15.83       & \cellcolor[HTML]{FFEB84}48\%  & 17.17       & \cellcolor[HTML]{FFE984}60\%  \\
\textbf{pt}        & 10.97          & 14.58       & \cellcolor[HTML]{C8DB80}33\%  & 14.60       & \cellcolor[HTML]{C9DB80}33\%  & 15.78       & \cellcolor[HTML]{F1E783}44\%  & 18.20       & \cellcolor[HTML]{FFE884}66\%  \\
\textbf{ar}        & 14.40          & 29.19       & \cellcolor[HTML]{FFE183}103\% & 26.27       & \cellcolor[HTML]{FFE583}82\%  & 22.82       & \cellcolor[HTML]{FFE984}58\%  & 17.65       & \cellcolor[HTML]{A2D07E}23\%  \\
\textbf{vi}        & 10.16          & 14.76       & \cellcolor[HTML]{F7E883}45\%  & 12.74       & \cellcolor[HTML]{ADD37F}25\%  & 13.48       & \cellcolor[HTML]{C8DB80}33\%  & 14.62       & \cellcolor[HTML]{F2E783}44\%  \\
\textbf{hi}        & 10.96          & 18.26       & \cellcolor[HTML]{FFE884}67\%  & 15.82       & \cellcolor[HTML]{F3E783}44\%  & 13.18       & \cellcolor[HTML]{99CD7E}20\%  & 13.97       & \cellcolor[HTML]{B4D57F}28\%  \\
\textbf{id}        & 20.48          & 29.37       & \cellcolor[HTML]{F0E683}43\%  & 27.30       & \cellcolor[HTML]{CADB80}33\%  & 27.48       & \cellcolor[HTML]{CDDC81}34\%  & 33.96       & \cellcolor[HTML]{FFE884}66\%  \\
\textbf{bn}        & 17.27          & 33.37       & \cellcolor[HTML]{FFE383}93\%  & 26.71       & \cellcolor[HTML]{FFEA84}55\%  & 21.94       & \cellcolor[HTML]{B2D57F}27\%  & 30.11       & \cellcolor[HTML]{FFE784}74\%  \\
\textbf{ta}        & 16.55          & 42.23       & \cellcolor[HTML]{FED881}155\% & 27.14       & \cellcolor[HTML]{FFE884}64\%  & 19.31       & \cellcolor[HTML]{8CC97D}17\%  & 24.44       & \cellcolor[HTML]{FFEB84}48\%  \\
\textbf{te}        & 18.10          & 64.97       & \cellcolor[HTML]{FDC57D}259\% & 39.77       & \cellcolor[HTML]{FFDE82}120\% & 25.58       & \cellcolor[HTML]{E8E482}41\%  & 37.68       & \cellcolor[HTML]{FFE082}108\% \\
\textbf{ur}        & 13.26          & 27.03       & \cellcolor[HTML]{FFE183}104\% & 25.07       & \cellcolor[HTML]{FFE483}89\%  & 19.70       & \cellcolor[HTML]{FFEB84}49\%  & 15.38       & \cellcolor[HTML]{89C97D}16\%  \\
\textbf{ne}        & 27.22          & 152.67      & \cellcolor[HTML]{FBA076}461\% & 64.74       & \cellcolor[HTML]{FFDB81}138\% & 46.59       & \cellcolor[HTML]{FFE784}71\%  & 46.60       & \cellcolor[HTML]{FFE784}71\%  \\
\textbf{mr}        & 23.07          & 176.78      & \cellcolor[HTML]{F97A6F}666\% & 66.86       & \cellcolor[HTML]{FED17F}190\% & 45.71       & \cellcolor[HTML]{FFE283}98\%  & 68.97       & \cellcolor[HTML]{FED07F}199\% \\
\textbf{gu}        & 21.52          & 184.62      & \cellcolor[HTML]{F8696B}758\% & 48.30       & \cellcolor[HTML]{FFDD82}124\% & 35.17       & \cellcolor[HTML]{FFE984}63\%  & 37.96       & \cellcolor[HTML]{FFE683}76\%  \\
\textbf{zh-Hant}   & 21.84          & 113.34      & \cellcolor[HTML]{FCA877}419\% & 49.41       & \cellcolor[HTML]{FFDD82}126\% & 56.20       & \cellcolor[HTML]{FED781}157\% & 36.81       & \cellcolor[HTML]{FFE884}69\%  \\
\textbf{sw}        & 34.35          & 145.54      & \cellcolor[HTML]{FDB97B}324\% & 58.74       & \cellcolor[HTML]{FFE784}71\%  & 91.94       & \cellcolor[HTML]{FED580}168\% & 125.09      & \cellcolor[HTML]{FDC47D}264\% \\
\textbf{yo}        & 53.29          & 128.12      & \cellcolor[HTML]{FFDA81}140\% & 72.86       & \cellcolor[HTML]{D7DF81}37\%  & 127.05      & \cellcolor[HTML]{FFDB81}138\% & 180.01      & \cellcolor[HTML]{FEC97E}238\% \\
\textbf{ig}        & 39.16          & 90.41       & \cellcolor[HTML]{FFDC82}131\% & 51.30       & \cellcolor[HTML]{C1D980}31\%  & 91.28       & \cellcolor[HTML]{FFDC81}133\% & 149.15      & \cellcolor[HTML]{FDC17C}281\% \\ \hline
\textbf{wikitext2} & 11.37          & 16.15       & \cellcolor[HTML]{EBE582}42\%  & 16.25       & \cellcolor[HTML]{EEE683}43\%  & 14.14       & \cellcolor[HTML]{A9D27F}24\%  & 15.68       & \cellcolor[HTML]{DBE081}38\%  \\  \hline
\textbf{Average}   & 20.08          & 64.70       & 189.1\%                       & 34.40       & 68.7\%                        & 36.49       & 63.4\%                        & 45.19       & 90.2\%      \\    \hline             
\end{tabular}
\caption{Monolingual pruning results using Wanda on BLOOM-7b1 with calibration data in \texttt{en}, \texttt{ig}, \texttt{ta}, and \texttt{ur}. Perplexity evaluated on XL-sum and wikitext2. Languages are ranked from the most well-represented to the least represented, from top to bottom.}
\end{table*}
\begin{table*}[]
\begin{tabular}{llllllllll}
\hline 
\textbf{Dataset}   & \textbf{Dense} & \textbf{en} & \textbf{}                     & \textbf{ig} & \textbf{}                    & \textbf{ta} & \textbf{}                     & \textbf{ur} & \textbf{}                     \\  \hline
\textbf{en}        & 13.68          & 14.92       & \cellcolor[HTML]{72C27B}9\%   & 16.28       & \cellcolor[HTML]{A0CF7E}19\% & 16.53       & \cellcolor[HTML]{A8D27F}21\%  & 16.80       & \cellcolor[HTML]{B1D47F}23\%  \\
\textbf{zh-Hans}   & 23.70          & 35.56       & \cellcolor[HTML]{FFE984}50\%  & 30.31       & \cellcolor[HTML]{C8DB80}28\% & 31.45       & \cellcolor[HTML]{DFE182}33\%  & 33.05       & \cellcolor[HTML]{FEEA83}39\%  \\
\textbf{fr}        & 9.59           & 13.41       & \cellcolor[HTML]{FFEB84}40\%  & 11.42       & \cellcolor[HTML]{A0CF7E}19\% & 12.71       & \cellcolor[HTML]{DEE182}32\%  & 12.75       & \cellcolor[HTML]{DFE282}33\%  \\
\textbf{es}        & 10.71          & 13.75       & \cellcolor[HTML]{CBDC81}28\%  & 12.82       & \cellcolor[HTML]{A3D07E}20\% & 14.12       & \cellcolor[HTML]{DBE081}32\%  & 14.58       & \cellcolor[HTML]{EEE683}36\%  \\
\textbf{pt}        & 10.97          & 17.37       & \cellcolor[HTML]{FFE784}58\%  & 13.49       & \cellcolor[HTML]{B2D47F}23\% & 14.86       & \cellcolor[HTML]{EBE582}35\%  & 15.38       & \cellcolor[HTML]{FFEB84}40\%  \\
\textbf{ar}        & 14.40          & 25.33       & \cellcolor[HTML]{FFE283}76\%  & 18.54       & \cellcolor[HTML]{CDDC81}29\% & 19.31       & \cellcolor[HTML]{E5E382}34\%  & 16.87       & \cellcolor[HTML]{97CD7E}17\%  \\
\textbf{vi}        & 10.16          & 15.00       & \cellcolor[HTML]{FFE984}48\%  & 12.07       & \cellcolor[HTML]{9FCF7E}19\% & 13.26       & \cellcolor[HTML]{D5DE81}31\%  & 13.07       & \cellcolor[HTML]{CCDC81}29\%  \\
\textbf{hi}        & 10.96          & 19.09       & \cellcolor[HTML]{FFE283}74\%  & 14.44       & \cellcolor[HTML]{DAE081}32\% & 12.73       & \cellcolor[HTML]{93CB7D}16\%  & 12.46       & \cellcolor[HTML]{88C87D}14\%  \\
\textbf{id}        & 20.48          & 37.27       & \cellcolor[HTML]{FFE082}82\%  & 25.56       & \cellcolor[HTML]{BAD780}25\% & 27.34       & \cellcolor[HTML]{E2E282}33\%  & 26.84       & \cellcolor[HTML]{D7DF81}31\%  \\
\textbf{bn}        & 17.27          & 40.50       & \cellcolor[HTML]{FED280}134\% & 24.21       & \cellcolor[HTML]{FFEB84}40\% & 21.07       & \cellcolor[HTML]{ADD37F}22\%  & 21.85       & \cellcolor[HTML]{C2D980}26\%  \\
\textbf{ta}        & 16.55          & 44.10       & \cellcolor[HTML]{FECA7E}167\% & 26.36       & \cellcolor[HTML]{FFE683}59\% & 17.84       & \cellcolor[HTML]{6DC07B}8\%   & 22.58       & \cellcolor[HTML]{F0E683}36\%  \\
\textbf{te}        & 18.10          & 69.20       & \cellcolor[HTML]{FCAB78}282\% & 31.56       & \cellcolor[HTML]{FFE283}74\% & 24.12       & \cellcolor[HTML]{E1E282}33\%  & 28.66       & \cellcolor[HTML]{FFE784}58\%  \\
\textbf{ur}        & 13.26          & 30.56       & \cellcolor[HTML]{FED380}130\% & 19.20       & \cellcolor[HTML]{FFEA84}45\% & 17.32       & \cellcolor[HTML]{D5DF81}31\%  & 14.12       & \cellcolor[HTML]{66BF7B}7\%   \\
\textbf{ne}        & 27.22          & 148.00      & \cellcolor[HTML]{FA8070}444\% & 47.77       & \cellcolor[HTML]{FFE283}76\% & 42.62       & \cellcolor[HTML]{FFE784}57\%  & 40.12       & \cellcolor[HTML]{FFE984}47\%  \\
\textbf{mr}        & 23.07          & 144.65      & \cellcolor[HTML]{F8696B}527\% & 43.74       & \cellcolor[HTML]{FFDE82}90\% & 37.81       & \cellcolor[HTML]{FFE583}64\%  & 49.41       & \cellcolor[HTML]{FED881}114\% \\
\textbf{gu}        & 21.52          & 118.48      & \cellcolor[HTML]{FA7E6F}450\% & 39.39       & \cellcolor[HTML]{FFE082}83\% & 34.87       & \cellcolor[HTML]{FFE683}62\%  & 32.87       & \cellcolor[HTML]{FFE884}53\%  \\
\textbf{zh-Hant}   & 21.84          & 102.26      & \cellcolor[HTML]{FB9474}368\% & 30.51       & \cellcolor[HTML]{FFEB84}40\% & 50.00       & \cellcolor[HTML]{FED480}129\% & 43.85       & \cellcolor[HTML]{FFDB81}101\% \\
\textbf{sw}        & 34.35          & 135.84      & \cellcolor[HTML]{FCA777}295\% & 46.74       & \cellcolor[HTML]{EEE683}36\% & 72.81       & \cellcolor[HTML]{FED881}112\% & 72.94       & \cellcolor[HTML]{FED881}112\% \\
\textbf{yo}        & 53.29          & 126.54      & \cellcolor[HTML]{FED17F}137\% & 61.08       & \cellcolor[HTML]{8CC97D}15\% & 105.97      & \cellcolor[HTML]{FFDC81}99\%  & 113.09      & \cellcolor[HTML]{FED881}112\% \\
\textbf{ig}        & 39.16          & 90.89       & \cellcolor[HTML]{FED380}132\% & 41.37       & \cellcolor[HTML]{63BE7B}6\%  & 74.44       & \cellcolor[HTML]{FFDE82}90\%  & 86.14       & \cellcolor[HTML]{FED680}120\% \\ \hline
\textbf{wikitext2} & 11.37          & 13.91       & \cellcolor[HTML]{AFD47F}22\%  & 13.56       & \cellcolor[HTML]{A1D07E}19\% & 13.66       & \cellcolor[HTML]{A5D17E}20\%  & 13.95       & \cellcolor[HTML]{B1D47F}23\%  \\ \hline
\textbf{Average}   & 20.08          & 59.84       & 169.3\%                       & 27.64       & 37.8\%                       & 32.14       & 47.3\%                        & 33.40       & 51.1\%                   \\    \hline
\end{tabular}
\caption{Monolingual pruning results using SparseGPT on BLOOM-7b1 with calibration data in \texttt{en}, \texttt{ig}, \texttt{ta}, and \texttt{ur}. Perplexity evaluated on XL-sum and wikitext2. Languages are ranked from the most well-represented to the least represented, from top to bottom.}
\end{table*}

\begin{table*}[]
\begin{tabular}{llllllllll}
\hline
\textbf{Dataset}   & \textbf{Dense} & \textbf{en} & \textbf{}                      & \textbf{ig} & \textbf{}                      & \textbf{te} & \textbf{}                      & \textbf{id} & \textbf{}                      \\ \hline
\textbf{en}        & 25.97        & 34.30       & \cellcolor[HTML]{63BE7B}32\%  & 49.17       & \cellcolor[HTML]{B6D57F}89\%  & 53.12       & \cellcolor[HTML]{CCDC81}105\%  & 38.50       & \cellcolor[HTML]{7AC47C}48\%  \\ 
\textbf{zh-Hans}   & 46.14        & 72.92       & \cellcolor[HTML]{88C87D}58\%  & 87.98       & \cellcolor[HTML]{B8D67F}91\%  & 126.56      & \cellcolor[HTML]{FFE984}174\%  & 79.06       & \cellcolor[HTML]{9BCE7E}71\%  \\ 
\textbf{zh-Hant}   & 43.79        & 137.00      & \cellcolor[HTML]{FFE784}213\% & 156.15      & \cellcolor[HTML]{FFE483}257\% & 227.56      & \cellcolor[HTML]{FFDA81}420\%  & 167.85      & \cellcolor[HTML]{FFE283}283\% \\ 
\textbf{fr}        & 16.32        & 22.89       & \cellcolor[HTML]{6EC17B}40\%  & 32.46       & \cellcolor[HTML]{C3D980}99\%  & 41.75       & \cellcolor[HTML]{FFEA84}156\%  & 24.91       & \cellcolor[HTML]{80C67C}53\%  \\ 
\textbf{es}        & 18.11        & 25.41       & \cellcolor[HTML]{6EC17B}40\%  & 38.29       & \cellcolor[HTML]{D6DF81}111\% & 47.38       & \cellcolor[HTML]{FFEA84}162\%  & 28.18       & \cellcolor[HTML]{85C77C}56\%  \\ 
\textbf{pt}        & 19.12        & 27.47       & \cellcolor[HTML]{73C27B}44\%  & 42.04       & \cellcolor[HTML]{E2E282}120\% & 57.68       & \cellcolor[HTML]{FFE884}202\%  & 30.47       & \cellcolor[HTML]{8AC97D}59\%  \\ 
\textbf{ar}        & 28.13        & 80.09       & \cellcolor[HTML]{FFE984}185\% & 79.26       & \cellcolor[HTML]{FFE984}182\% & 66.66       & \cellcolor[HTML]{FBE983}137\%  & 55.03       & \cellcolor[HTML]{BFD880}96\%  \\ 
\textbf{vi}        & 20.71        & 43.26       & \cellcolor[HTML]{D2DE81}109\% & 39.63       & \cellcolor[HTML]{B9D67F}91\%  & 150.26      & \cellcolor[HTML]{FECC7F}625\%  & 38.05       & \cellcolor[HTML]{ADD37F}84\%  \\ 
\textbf{hi}        & 20.86        & 44.10       & \cellcolor[HTML]{D6DF81}111\% & 50.54       & \cellcolor[HTML]{FFEB84}142\% & 32.37       & \cellcolor[HTML]{84C77C}55\%   & 36.25       & \cellcolor[HTML]{9FCF7E}74\%  \\ 
\textbf{id}        & 42.05        & 109.13      & \cellcolor[HTML]{FFEA84}160\% & 108.26      & \cellcolor[HTML]{FFEA84}157\% & 952.66      & \cellcolor[HTML]{F8696B}2166\% & 60.77       & \cellcolor[HTML]{75C37C}45\%  \\ 
\textbf{bn}        & 37.05        & 102.78      & \cellcolor[HTML]{FFE984}177\% & 141.25      & \cellcolor[HTML]{FFE283}281\% & 68.01       & \cellcolor[HTML]{ADD37F}84\%   & 83.81       & \cellcolor[HTML]{EBE582}126\% \\ 
\textbf{ta}        & 36.97        & 176.25      & \cellcolor[HTML]{FFDC82}377\% & 147.99      & \cellcolor[HTML]{FFE183}300\% & 60.52       & \cellcolor[HTML]{90CB7D}64\%   & 163.29      & \cellcolor[HTML]{FFDF82}342\% \\ 
\textbf{te}        & 54.40        & 355.98      & \cellcolor[HTML]{FED17F}554\% & 260.31      & \cellcolor[HTML]{FFDC82}379\% & 95.77       & \cellcolor[HTML]{A2D07E}76\%   & 445.67      & \cellcolor[HTML]{FDC67D}719\% \\ 
\textbf{ur}        & 29.42        & 97.65       & \cellcolor[HTML]{FFE683}232\% & 143.11      & \cellcolor[HTML]{FFDC81}386\% & 49.14       & \cellcolor[HTML]{95CC7D}67\%   & 63.39       & \cellcolor[HTML]{DBE081}115\% \\ 
\textbf{ne}        & 75.80        & 555.21      & \cellcolor[HTML]{FECC7E}632\% & 320.61      & \cellcolor[HTML]{FFE082}323\% & 143.19      & \cellcolor[HTML]{B5D57F}89\%   & 237.41      & \cellcolor[HTML]{FFE784}213\% \\ 
\textbf{mr}        & 72.45        & 503.51      & \cellcolor[HTML]{FECE7F}595\% & 290.52      & \cellcolor[HTML]{FFE183}301\% & 140.02      & \cellcolor[HTML]{BBD780}93\%   & 220.62      & \cellcolor[HTML]{FFE784}205\% \\ 
\textbf{gu}        & 70.65        & 327.11      & \cellcolor[HTML]{FFDD82}363\% & 215.13      & \cellcolor[HTML]{FFE784}205\% & 124.55      & \cellcolor[HTML]{A3D07E}76\%   & 286.66      & \cellcolor[HTML]{FFE182}306\% \\ 
\textbf{sw}        & 154.76       & 919.14      & \cellcolor[HTML]{FED580}494\% & 419.52      & \cellcolor[HTML]{FFE984}171\% & 897.07      & \cellcolor[HTML]{FED680}480\%  & 465.66      & \cellcolor[HTML]{FFE884}201\% \\ 
\textbf{yo}        & 267.02       & 1024.94     & \cellcolor[HTML]{FFE283}284\% & 529.35      & \cellcolor[HTML]{C3D980}98\%  & 711.20      & \cellcolor[HTML]{FFEA84}166\%  & 573.17      & \cellcolor[HTML]{DAE081}115\% \\ 
\textbf{ig}        & 185.74       & 728.01      & \cellcolor[HTML]{FFE283}292\% & 285.28      & \cellcolor[HTML]{82C77C}54\%  & 773.60      & \cellcolor[HTML]{FFE082}316\%  & 409.11      & \cellcolor[HTML]{E2E282}120\% \\ 
\textbf{wikitext2} & 22.41        & 30.58       & \cellcolor[HTML]{69BF7B}36\%  & 40.70       & \cellcolor[HTML]{AAD27F}82\%  & 46.07       & \cellcolor[HTML]{CDDC81}106\%  & 33.40       & \cellcolor[HTML]{7BC57C}49\%  \\ \hline
\textbf{Average}       & 61.33        & 257.99      & 239.5\%                       & 165.60      & 186.6\%                       & 231.67      & 277.0\%                        & 168.63      & 160.9\%                       \\ \hline
\end{tabular}

\caption{Monolingual pruning results of Wanda on BLOOM-560m.}
\end{table*}

\begin{table*}[]
\centering

\begin{tabular}{llllll}
\hline
\textbf{Dataset}   & \textbf{Dense} & \textbf{en}                      & \textbf{ig}                      & \textbf{te}                      & \textbf{id}                      \\ \hline
\textbf{en}        & 25.97          & \cellcolor[HTML]{64BE7B}32.02    & \cellcolor[HTML]{66BF7B}43.04    & \cellcolor[HTML]{67BF7B}48.17    & \cellcolor[HTML]{66BE7B}39.53    \\ 
\textbf{zh-Hans}   & 46.14          & \cellcolor[HTML]{72C27B}101.79   & \cellcolor[HTML]{78C47C}127.73   & \cellcolor[HTML]{74C27B}107.62   & \cellcolor[HTML]{74C37C}109.50   \\ 
\textbf{zh-Hant}   & 43.79          & \cellcolor[HTML]{94CC7D}266.49   & \cellcolor[HTML]{7FC67C}161.94   & \cellcolor[HTML]{8CCA7D}227.24   & \cellcolor[HTML]{97CD7E}279.39   \\
\textbf{fr}        & 16.32          & \cellcolor[HTML]{63BE7B}24.99    & \cellcolor[HTML]{63BE7B}28.45    & \cellcolor[HTML]{65BE7B}38.85    & \cellcolor[HTML]{63BE7B}27.65    \\ 
\textbf{es}        & 18.11          & \cellcolor[HTML]{63BE7B}27.94    & \cellcolor[HTML]{64BE7B}34.01    & \cellcolor[HTML]{67BF7B}47.80    & \cellcolor[HTML]{64BE7B}32.46    \\ 
\textbf{pt}        & 19.12          & \cellcolor[HTML]{65BE7B}34.99    & \cellcolor[HTML]{65BE7B}36.88    & \cellcolor[HTML]{69BF7B}54.17    & \cellcolor[HTML]{64BE7B}34.60    \\ 
\textbf{ar}        & 28.13          & \cellcolor[HTML]{FFEB84}1.13E+14 & \cellcolor[HTML]{FFEB84}8.10E+07 & \cellcolor[HTML]{90CB7D}246.62   & \cellcolor[HTML]{FFEB84}3.52E+07 \\ 
\textbf{vi}        & 20.71          & \cellcolor[HTML]{78C47C}127.59   & \cellcolor[HTML]{6CC07B}69.64    & \cellcolor[HTML]{FFEB84}947.65   & \cellcolor[HTML]{6DC17B}77.75    \\ 
\textbf{hi}        & 20.86          & \cellcolor[HTML]{FFEB84}1.20E+15 & \cellcolor[HTML]{FFEB84}3.93E+10 & \cellcolor[HTML]{78C47C}128.37   & \cellcolor[HTML]{FFEB84}2.98E+09 \\
\textbf{id}        & 42.05          & \cellcolor[HTML]{FFEB84}76038.83 & \cellcolor[HTML]{FFEB84}7049.31  & \cellcolor[HTML]{FFEB84}1.14E+04 & \cellcolor[HTML]{6AC07B}62.01    \\ 
\textbf{bn}        & 37.05          & \cellcolor[HTML]{F8696B}4.42E+23 & \cellcolor[HTML]{FFEB84}2.16E+10 & \cellcolor[HTML]{FFEB84}1787.97  & \cellcolor[HTML]{FFEB84}6.12E+15 \\ 
\textbf{ta}        & 36.97          & \cellcolor[HTML]{FFEB84}1.71E+07 & \cellcolor[HTML]{FFEB84}3.61E+05 & \cellcolor[HTML]{6FC17B}84.92    & \cellcolor[HTML]{FFEB84}1.80E+08 \\ 
\textbf{te}        & 54.40          & \cellcolor[HTML]{FFEB84}6.96E+05 & \cellcolor[HTML]{FFEB84}1.80E+05 & \cellcolor[HTML]{6DC07B}74.68    & \cellcolor[HTML]{FFEB84}2.40E+07 \\ 
\textbf{ur}        & 29.42          & \cellcolor[HTML]{FFEB84}3.62E+20 & \cellcolor[HTML]{FFEB84}5.53E+04 & \cellcolor[HTML]{FFEB84}807.08   & \cellcolor[HTML]{FFEB84}1.55E+07 \\ 
\textbf{ne}        & 75.80          & \cellcolor[HTML]{FFEB84}1.27E+17 & \cellcolor[HTML]{FFEB84}1.70E+09 & \cellcolor[HTML]{E1E282}635.82   & \cellcolor[HTML]{FFEB84}6.43E+10 \\ 
\textbf{mr}        & 72.45          & \cellcolor[HTML]{FFEB84}4.25E+12 & \cellcolor[HTML]{FFEB84}2.13E+08 & \cellcolor[HTML]{FFEB84}1249.71  & \cellcolor[HTML]{FFEB84}3.76E+11 \\ 
\textbf{gu}        & 70.65          & \cellcolor[HTML]{FFEB84}2.17E+13 & \cellcolor[HTML]{FFEB84}3050.59  & \cellcolor[HTML]{91CB7D}252.15   & \cellcolor[HTML]{FFEB84}4.96E+06 \\ 
\textbf{sw}        & 154.76         & \cellcolor[HTML]{FFEB84}6011.36  & \cellcolor[HTML]{BED880}466.51   & \cellcolor[HTML]{FFEB84}1363.16  & \cellcolor[HTML]{FFEB84}1502.12  \\ 
\textbf{yo}        & 267.02         & \cellcolor[HTML]{FFEB84}3557.88  & \cellcolor[HTML]{A4D07E}342.02   & \cellcolor[HTML]{FFEB84}881.53   & \cellcolor[HTML]{FFEB84}1462.66  \\ 
\textbf{ig}        & 185.74         & \cellcolor[HTML]{FFEB84}1306.30  & \cellcolor[HTML]{88C87D}204.65   & \cellcolor[HTML]{FFEB84}989.57   & \cellcolor[HTML]{F9E983}753.18   \\ 
\textbf{wikitext2} & 22.41          & \cellcolor[HTML]{63BE7B}29.75    & \cellcolor[HTML]{65BE7B}36.37    & \cellcolor[HTML]{66BE7B}40.72    & \cellcolor[HTML]{64BE7B}34.33    \\ \hline
\textbf{Average}   & 61.33          & 2.11E+22                         & 2.99E+09                         & 1020                             & 2.92E+14                         \\ \hline
\end{tabular}

\caption{Monolingual pruning results of SparseGPT on BLOOM-560m.}
\end{table*}

\begin{table*}[]
\centering
\tiny
\resizebox{\textwidth}{!}{
\begin{tabular}{lllllllllllllllllllll}
                 & \textbf{en}                & \textbf{zh-Hans}           & \textbf{zh-Hant}           & \textbf{fr}                & \textbf{es}                & \textbf{pt}                & \textbf{ar}                & \textbf{vi}                & \textbf{hi}                & \textbf{id}                & \textbf{bn}                & \textbf{ta}                & \textbf{te}                & \textbf{ur}                & \textbf{ne}                & \textbf{mr}                & \textbf{gu}                & \textbf{sw}                & \textbf{yo}                & \textbf{ig}                \\
\textbf{en}      & \cellcolor[HTML]{63BE7B}0  & \cellcolor[HTML]{FFE383}10 & \cellcolor[HTML]{FFDB81}11 & \cellcolor[HTML]{B9D780}5  & \cellcolor[HTML]{CBDC81}6  & \cellcolor[HTML]{DCE182}7  & \cellcolor[HTML]{FFEB84}9  & \cellcolor[HTML]{EDE683}8  & \cellcolor[HTML]{FFE383}10 & \cellcolor[HTML]{CBDC81}6  & \cellcolor[HTML]{FFEB84}9  & \cellcolor[HTML]{EDE683}8  & \cellcolor[HTML]{FED380}12 & \cellcolor[HTML]{FB9273}20 & \cellcolor[HTML]{FDC37D}14 & \cellcolor[HTML]{FFEB84}9  & \cellcolor[HTML]{FED380}12 & \cellcolor[HTML]{FFE383}10 & \cellcolor[HTML]{FFE383}10 & \cellcolor[HTML]{FFE383}10 \\
\textbf{zh-Hans} & \cellcolor[HTML]{FFE383}10 & \cellcolor[HTML]{63BE7B}0  & \cellcolor[HTML]{74C37C}1  & \cellcolor[HTML]{FECB7E}13 & \cellcolor[HTML]{FDBB7B}15 & \cellcolor[HTML]{FCB37A}16 & \cellcolor[HTML]{EDE683}8  & \cellcolor[HTML]{FED380}12 & \cellcolor[HTML]{97CD7E}3  & \cellcolor[HTML]{FED380}12 & \cellcolor[HTML]{CBDC81}6  & \cellcolor[HTML]{CBDC81}6  & \cellcolor[HTML]{B9D780}5  & \cellcolor[HTML]{FFDB81}11 & \cellcolor[HTML]{EDE683}8  & \cellcolor[HTML]{B9D780}5  & \cellcolor[HTML]{CBDC81}6  & \cellcolor[HTML]{FCB37A}16 & \cellcolor[HTML]{FCAA78}17 & \cellcolor[HTML]{FCA276}18 \\
\textbf{zh-Hant} & \cellcolor[HTML]{FFDB81}11 & \cellcolor[HTML]{74C37C}1  & \cellcolor[HTML]{63BE7B}0  & \cellcolor[HTML]{FDBB7B}15 & \cellcolor[HTML]{FCAA78}17 & \cellcolor[HTML]{FCA276}18 & \cellcolor[HTML]{EDE683}8  & \cellcolor[HTML]{FDC37D}14 & \cellcolor[HTML]{A8D27F}4  & \cellcolor[HTML]{FECB7E}13 & \cellcolor[HTML]{CBDC81}6  & \cellcolor[HTML]{CBDC81}6  & \cellcolor[HTML]{B9D780}5  & \cellcolor[HTML]{FFE383}10 & \cellcolor[HTML]{DCE182}7  & \cellcolor[HTML]{CBDC81}6  & \cellcolor[HTML]{CBDC81}6  & \cellcolor[HTML]{FCAA78}17 & \cellcolor[HTML]{FB9A75}19 & \cellcolor[HTML]{FB9A75}19 \\
\textbf{fr}      & \cellcolor[HTML]{B9D780}5  & \cellcolor[HTML]{FECB7E}13 & \cellcolor[HTML]{FDBB7B}15 & \cellcolor[HTML]{63BE7B}0  & \cellcolor[HTML]{85C87D}2  & \cellcolor[HTML]{97CD7E}3  & \cellcolor[HTML]{FFDB81}11 & \cellcolor[HTML]{B9D780}5  & \cellcolor[HTML]{FED380}12 & \cellcolor[HTML]{A8D27F}4  & \cellcolor[HTML]{FFE383}10 & \cellcolor[HTML]{FFE383}10 & \cellcolor[HTML]{FDC37D}14 & \cellcolor[HTML]{FA8270}22 & \cellcolor[HTML]{FCB37A}16 & \cellcolor[HTML]{FFDB81}11 & \cellcolor[HTML]{FECB7E}13 & \cellcolor[HTML]{CBDC81}6  & \cellcolor[HTML]{CBDC81}6  & \cellcolor[HTML]{DCE182}7  \\
\textbf{es}      & \cellcolor[HTML]{CBDC81}6  & \cellcolor[HTML]{FDBB7B}15 & \cellcolor[HTML]{FCAA78}17 & \cellcolor[HTML]{85C87D}2  & \cellcolor[HTML]{63BE7B}0  & \cellcolor[HTML]{74C37C}1  & \cellcolor[HTML]{FED380}12 & \cellcolor[HTML]{DCE182}7  & \cellcolor[HTML]{FDBB7B}15 & \cellcolor[HTML]{CBDC81}6  & \cellcolor[HTML]{FECB7E}13 & \cellcolor[HTML]{FED380}12 & \cellcolor[HTML]{FCAA78}17 & \cellcolor[HTML]{F8696B}25 & \cellcolor[HTML]{FCA276}18 & \cellcolor[HTML]{FECB7E}13 & \cellcolor[HTML]{FDBB7B}15 & \cellcolor[HTML]{DCE182}7  & \cellcolor[HTML]{B9D780}5  & \cellcolor[HTML]{CBDC81}6  \\
\textbf{pt}      & \cellcolor[HTML]{DCE182}7  & \cellcolor[HTML]{FCB37A}16 & \cellcolor[HTML]{FCA276}18 & \cellcolor[HTML]{97CD7E}3  & \cellcolor[HTML]{74C37C}1  & \cellcolor[HTML]{63BE7B}0  & \cellcolor[HTML]{FED380}12 & \cellcolor[HTML]{DCE182}7  & \cellcolor[HTML]{FDBB7B}15 & \cellcolor[HTML]{B9D780}5  & \cellcolor[HTML]{FECB7E}13 & \cellcolor[HTML]{FECB7E}13 & \cellcolor[HTML]{FCAA78}17 & \cellcolor[HTML]{F8696B}25 & \cellcolor[HTML]{FCA276}18 & \cellcolor[HTML]{FECB7E}13 & \cellcolor[HTML]{FCB37A}16 & \cellcolor[HTML]{B9D780}5  & \cellcolor[HTML]{A8D27F}4  & \cellcolor[HTML]{B9D780}5  \\
\textbf{ar}      & \cellcolor[HTML]{FFEB84}9  & \cellcolor[HTML]{EDE683}8  & \cellcolor[HTML]{EDE683}8  & \cellcolor[HTML]{FFDB81}11 & \cellcolor[HTML]{FED380}12 & \cellcolor[HTML]{FED380}12 & \cellcolor[HTML]{63BE7B}0  & \cellcolor[HTML]{EDE683}8  & \cellcolor[HTML]{CBDC81}6  & \cellcolor[HTML]{EDE683}8  & \cellcolor[HTML]{CBDC81}6  & \cellcolor[HTML]{B9D780}5  & \cellcolor[HTML]{EDE683}8  & \cellcolor[HTML]{FDC37D}14 & \cellcolor[HTML]{EDE683}8  & \cellcolor[HTML]{A8D27F}4  & \cellcolor[HTML]{EDE683}8  & \cellcolor[HTML]{FFDB81}11 & \cellcolor[HTML]{FED380}12 & \cellcolor[HTML]{FED380}12 \\
\textbf{vi}      & \cellcolor[HTML]{EDE683}8  & \cellcolor[HTML]{FED380}12 & \cellcolor[HTML]{FDC37D}14 & \cellcolor[HTML]{B9D780}5  & \cellcolor[HTML]{DCE182}7  & \cellcolor[HTML]{DCE182}7  & \cellcolor[HTML]{EDE683}8  & \cellcolor[HTML]{63BE7B}0  & \cellcolor[HTML]{FFE383}10 & \cellcolor[HTML]{85C87D}2  & \cellcolor[HTML]{EDE683}8  & \cellcolor[HTML]{EDE683}8  & \cellcolor[HTML]{FED380}12 & \cellcolor[HTML]{FB9A75}19 & \cellcolor[HTML]{FED380}12 & \cellcolor[HTML]{EDE683}8  & \cellcolor[HTML]{FFE383}10 & \cellcolor[HTML]{A8D27F}4  & \cellcolor[HTML]{CBDC81}6  & \cellcolor[HTML]{CBDC81}6  \\
\textbf{hi}      & \cellcolor[HTML]{FFE383}10 & \cellcolor[HTML]{97CD7E}3  & \cellcolor[HTML]{A8D27F}4  & \cellcolor[HTML]{FED380}12 & \cellcolor[HTML]{FDBB7B}15 & \cellcolor[HTML]{FDBB7B}15 & \cellcolor[HTML]{CBDC81}6  & \cellcolor[HTML]{FFE383}10 & \cellcolor[HTML]{63BE7B}0  & \cellcolor[HTML]{FFE383}10 & \cellcolor[HTML]{97CD7E}3  & \cellcolor[HTML]{97CD7E}3  & \cellcolor[HTML]{97CD7E}3  & \cellcolor[HTML]{FFE383}10 & \cellcolor[HTML]{B9D780}5  & \cellcolor[HTML]{97CD7E}3  & \cellcolor[HTML]{A8D27F}4  & \cellcolor[HTML]{FDC37D}14 & \cellcolor[HTML]{FCB37A}16 & \cellcolor[HTML]{FCB37A}16 \\
\textbf{id}      & \cellcolor[HTML]{CBDC81}6  & \cellcolor[HTML]{FED380}12 & \cellcolor[HTML]{FECB7E}13 & \cellcolor[HTML]{A8D27F}4  & \cellcolor[HTML]{CBDC81}6  & \cellcolor[HTML]{B9D780}5  & \cellcolor[HTML]{EDE683}8  & \cellcolor[HTML]{85C87D}2  & \cellcolor[HTML]{FFE383}10 & \cellcolor[HTML]{63BE7B}0  & \cellcolor[HTML]{EDE683}8  & \cellcolor[HTML]{DCE182}7  & \cellcolor[HTML]{FED380}12 & \cellcolor[HTML]{FB9A75}19 & \cellcolor[HTML]{FECB7E}13 & \cellcolor[HTML]{EDE683}8  & \cellcolor[HTML]{FFE383}10 & \cellcolor[HTML]{A8D27F}4  & \cellcolor[HTML]{CBDC81}6  & \cellcolor[HTML]{CBDC81}6  \\
\textbf{bn}      & \cellcolor[HTML]{FFEB84}9  & \cellcolor[HTML]{CBDC81}6  & \cellcolor[HTML]{CBDC81}6  & \cellcolor[HTML]{FFE383}10 & \cellcolor[HTML]{FECB7E}13 & \cellcolor[HTML]{FECB7E}13 & \cellcolor[HTML]{CBDC81}6  & \cellcolor[HTML]{EDE683}8  & \cellcolor[HTML]{97CD7E}3  & \cellcolor[HTML]{EDE683}8  & \cellcolor[HTML]{63BE7B}0  & \cellcolor[HTML]{85C87D}2  & \cellcolor[HTML]{A8D27F}4  & \cellcolor[HTML]{FED380}12 & \cellcolor[HTML]{CBDC81}6  & \cellcolor[HTML]{97CD7E}3  & \cellcolor[HTML]{97CD7E}3  & \cellcolor[HTML]{FED380}12 & \cellcolor[HTML]{FDC37D}14 & \cellcolor[HTML]{FDC37D}14 \\
\textbf{ta}      & \cellcolor[HTML]{EDE683}8  & \cellcolor[HTML]{CBDC81}6  & \cellcolor[HTML]{CBDC81}6  & \cellcolor[HTML]{FFE383}10 & \cellcolor[HTML]{FED380}12 & \cellcolor[HTML]{FECB7E}13 & \cellcolor[HTML]{B9D780}5  & \cellcolor[HTML]{EDE683}8  & \cellcolor[HTML]{97CD7E}3  & \cellcolor[HTML]{DCE182}7  & \cellcolor[HTML]{85C87D}2  & \cellcolor[HTML]{63BE7B}0  & \cellcolor[HTML]{A8D27F}4  & \cellcolor[HTML]{FED380}12 & \cellcolor[HTML]{CBDC81}6  & \cellcolor[HTML]{85C87D}2  & \cellcolor[HTML]{A8D27F}4  & \cellcolor[HTML]{FFDB81}11 & \cellcolor[HTML]{FECB7E}13 & \cellcolor[HTML]{FECB7E}13 \\
\textbf{te}      & \cellcolor[HTML]{FED380}12 & \cellcolor[HTML]{B9D780}5  & \cellcolor[HTML]{B9D780}5  & \cellcolor[HTML]{FDC37D}14 & \cellcolor[HTML]{FCAA78}17 & \cellcolor[HTML]{FCAA78}17 & \cellcolor[HTML]{EDE683}8  & \cellcolor[HTML]{FED380}12 & \cellcolor[HTML]{97CD7E}3  & \cellcolor[HTML]{FED380}12 & \cellcolor[HTML]{A8D27F}4  & \cellcolor[HTML]{A8D27F}4  & \cellcolor[HTML]{63BE7B}0  & \cellcolor[HTML]{EDE683}8  & \cellcolor[HTML]{A8D27F}4  & \cellcolor[HTML]{B9D780}5  & \cellcolor[HTML]{85C87D}2  & \cellcolor[HTML]{FCB37A}16 & \cellcolor[HTML]{FCA276}18 & \cellcolor[HTML]{FCA276}18 \\
\textbf{ur}      & \cellcolor[HTML]{FB9273}20 & \cellcolor[HTML]{FFDB81}11 & \cellcolor[HTML]{FFE383}10 & \cellcolor[HTML]{FA8270}22 & \cellcolor[HTML]{F8696B}25 & \cellcolor[HTML]{F8696B}25 & \cellcolor[HTML]{FDC37D}14 & \cellcolor[HTML]{FB9A75}19 & \cellcolor[HTML]{FFE383}10 & \cellcolor[HTML]{FB9A75}19 & \cellcolor[HTML]{FED380}12 & \cellcolor[HTML]{FED380}12 & \cellcolor[HTML]{EDE683}8  & \cellcolor[HTML]{63BE7B}0  & \cellcolor[HTML]{DCE182}7  & \cellcolor[HTML]{FED380}12 & \cellcolor[HTML]{FFE383}10 & \cellcolor[HTML]{F97A6F}23 & \cellcolor[HTML]{F8696B}25 & \cellcolor[HTML]{F8696B}25 \\
\textbf{ne}      & \cellcolor[HTML]{FDC37D}14 & \cellcolor[HTML]{EDE683}8  & \cellcolor[HTML]{DCE182}7  & \cellcolor[HTML]{FCB37A}16 & \cellcolor[HTML]{FCA276}18 & \cellcolor[HTML]{FCA276}18 & \cellcolor[HTML]{EDE683}8  & \cellcolor[HTML]{FED380}12 & \cellcolor[HTML]{B9D780}5  & \cellcolor[HTML]{FECB7E}13 & \cellcolor[HTML]{CBDC81}6  & \cellcolor[HTML]{CBDC81}6  & \cellcolor[HTML]{A8D27F}4  & \cellcolor[HTML]{DCE182}7  & \cellcolor[HTML]{63BE7B}0  & \cellcolor[HTML]{B9D780}5  & \cellcolor[HTML]{B9D780}5  & \cellcolor[HTML]{FCB37A}16 & \cellcolor[HTML]{FCA276}18 & \cellcolor[HTML]{FCA276}18 \\
\textbf{mr}      & \cellcolor[HTML]{FFEB84}9  & \cellcolor[HTML]{B9D780}5  & \cellcolor[HTML]{CBDC81}6  & \cellcolor[HTML]{FFDB81}11 & \cellcolor[HTML]{FECB7E}13 & \cellcolor[HTML]{FECB7E}13 & \cellcolor[HTML]{A8D27F}4  & \cellcolor[HTML]{EDE683}8  & \cellcolor[HTML]{97CD7E}3  & \cellcolor[HTML]{EDE683}8  & \cellcolor[HTML]{97CD7E}3  & \cellcolor[HTML]{85C87D}2  & \cellcolor[HTML]{B9D780}5  & \cellcolor[HTML]{FED380}12 & \cellcolor[HTML]{B9D780}5  & \cellcolor[HTML]{63BE7B}0  & \cellcolor[HTML]{A8D27F}4  & \cellcolor[HTML]{FED380}12 & \cellcolor[HTML]{FECB7E}13 & \cellcolor[HTML]{FECB7E}13 \\
\textbf{gu}      & \cellcolor[HTML]{FED380}12 & \cellcolor[HTML]{CBDC81}6  & \cellcolor[HTML]{CBDC81}6  & \cellcolor[HTML]{FECB7E}13 & \cellcolor[HTML]{FDBB7B}15 & \cellcolor[HTML]{FCB37A}16 & \cellcolor[HTML]{EDE683}8  & \cellcolor[HTML]{FFE383}10 & \cellcolor[HTML]{A8D27F}4  & \cellcolor[HTML]{FFE383}10 & \cellcolor[HTML]{97CD7E}3  & \cellcolor[HTML]{A8D27F}4  & \cellcolor[HTML]{85C87D}2  & \cellcolor[HTML]{FFE383}10 & \cellcolor[HTML]{B9D780}5  & \cellcolor[HTML]{A8D27F}4  & \cellcolor[HTML]{63BE7B}0  & \cellcolor[HTML]{FDC37D}14 & \cellcolor[HTML]{FCB37A}16 & \cellcolor[HTML]{FCAA78}17 \\
\textbf{sw}      & \cellcolor[HTML]{FFE383}10 & \cellcolor[HTML]{FCB37A}16 & \cellcolor[HTML]{FCAA78}17 & \cellcolor[HTML]{CBDC81}6  & \cellcolor[HTML]{DCE182}7  & \cellcolor[HTML]{B9D780}5  & \cellcolor[HTML]{FFDB81}11 & \cellcolor[HTML]{A8D27F}4  & \cellcolor[HTML]{FDC37D}14 & \cellcolor[HTML]{A8D27F}4  & \cellcolor[HTML]{FED380}12 & \cellcolor[HTML]{FFDB81}11 & \cellcolor[HTML]{FCB37A}16 & \cellcolor[HTML]{F97A6F}23 & \cellcolor[HTML]{FCB37A}16 & \cellcolor[HTML]{FED380}12 & \cellcolor[HTML]{FDC37D}14 & \cellcolor[HTML]{63BE7B}0  & \cellcolor[HTML]{97CD7E}3  & \cellcolor[HTML]{97CD7E}3  \\
\textbf{yo}      & \cellcolor[HTML]{FFE383}10 & \cellcolor[HTML]{FCAA78}17 & \cellcolor[HTML]{FB9A75}19 & \cellcolor[HTML]{CBDC81}6  & \cellcolor[HTML]{B9D780}5  & \cellcolor[HTML]{A8D27F}4  & \cellcolor[HTML]{FED380}12 & \cellcolor[HTML]{CBDC81}6  & \cellcolor[HTML]{FCB37A}16 & \cellcolor[HTML]{CBDC81}6  & \cellcolor[HTML]{FDC37D}14 & \cellcolor[HTML]{FECB7E}13 & \cellcolor[HTML]{FCA276}18 & \cellcolor[HTML]{F8696B}25 & \cellcolor[HTML]{FCA276}18 & \cellcolor[HTML]{FECB7E}13 & \cellcolor[HTML]{FCB37A}16 & \cellcolor[HTML]{97CD7E}3  & \cellcolor[HTML]{63BE7B}0  & \cellcolor[HTML]{74C37C}1  \\
\textbf{ig}      & \cellcolor[HTML]{FFE383}10 & \cellcolor[HTML]{FCA276}18 & \cellcolor[HTML]{FB9A75}19 & \cellcolor[HTML]{DCE182}7  & \cellcolor[HTML]{CBDC81}6  & \cellcolor[HTML]{B9D780}5  & \cellcolor[HTML]{FED380}12 & \cellcolor[HTML]{CBDC81}6  & \cellcolor[HTML]{FCB37A}16 & \cellcolor[HTML]{CBDC81}6  & \cellcolor[HTML]{FDC37D}14 & \cellcolor[HTML]{FECB7E}13 & \cellcolor[HTML]{FCA276}18 & \cellcolor[HTML]{F8696B}25 & \cellcolor[HTML]{FCA276}18 & \cellcolor[HTML]{FECB7E}13 & \cellcolor[HTML]{FCAA78}17 & \cellcolor[HTML]{97CD7E}3  & \cellcolor[HTML]{74C37C}1  & \cellcolor[HTML]{63BE7B}0  \\
\textbf{Average} & 9.3                        & 9.4                        & 10.1                       & 9.25                       & 10.6                       & 10.65                      & 8.5                        & 8.3                        & 8.1                        & 7.95                       & 7.6                        & 7.25                       & 9.2                        & 15.45                      & 10.2                       & 7.45                       & 8.75                       & 10.2                       & 11.1                       & 11.35                     
\end{tabular}}

\caption{Original distance matrix generated from the BLOOM-7b1 model.}
\label{dist7b1}
\end{table*}

\begin{table*}[]
\tiny
\resizebox{\textwidth}{!}{
\begin{tabular}{lllllllllllllllllllll}
                 & \textbf{en}                & \textbf{zh-Hans}           & \textbf{zh-Hant}           & \textbf{fr}                & \textbf{es}                & \textbf{pt}                & \textbf{ar}                & \textbf{vi}                & \textbf{hi}                & \textbf{id}                & \textbf{bn}                & \textbf{ta}                & \textbf{te}                & \textbf{ur}                & \textbf{ne}                & \textbf{mr}                & \textbf{gu}                & \textbf{sw}                & \textbf{yo}                & \textbf{ig}                \\
\textbf{en}      & \cellcolor[HTML]{63BE7B}0  & \cellcolor[HTML]{94CC7D}8  & \cellcolor[HTML]{9BCE7E}9  & \cellcolor[HTML]{6FC17B}2  & \cellcolor[HTML]{69BF7B}1  & \cellcolor[HTML]{7BC57C}4  & \cellcolor[HTML]{FA8070}54 & \cellcolor[HTML]{BAD780}14 & \cellcolor[HTML]{FFE082}28 & \cellcolor[HTML]{ADD37F}12 & \cellcolor[HTML]{FB9A75}47 & \cellcolor[HTML]{FB9674}48 & \cellcolor[HTML]{F8696B}60 & \cellcolor[HTML]{FA8771}52 & \cellcolor[HTML]{FDBB7B}38 & \cellcolor[HTML]{FED580}31 & \cellcolor[HTML]{F97C6F}55 & \cellcolor[HTML]{C0D980}15 & \cellcolor[HTML]{B4D57F}13 & \cellcolor[HTML]{B4D57F}13 \\
\textbf{zh-Hans} & \cellcolor[HTML]{94CC7D}8  & \cellcolor[HTML]{63BE7B}0  & \cellcolor[HTML]{69BF7B}1  & \cellcolor[HTML]{8ECA7D}7  & \cellcolor[HTML]{94CC7D}8  & \cellcolor[HTML]{82C77C}5  & \cellcolor[HTML]{FA8370}53 & \cellcolor[HTML]{8ECA7D}7  & \cellcolor[HTML]{F8E983}24 & \cellcolor[HTML]{88C87D}6  & \cellcolor[HTML]{FBA176}45 & \cellcolor[HTML]{FB9D75}46 & \cellcolor[HTML]{F9716D}58 & \cellcolor[HTML]{FB9273}49 & \cellcolor[HTML]{FECA7E}34 & \cellcolor[HTML]{FFE082}28 & \cellcolor[HTML]{FA8370}53 & \cellcolor[HTML]{8ECA7D}7  & \cellcolor[HTML]{82C77C}5  & \cellcolor[HTML]{82C77C}5  \\
\textbf{zh-Hant} & \cellcolor[HTML]{9BCE7E}9  & \cellcolor[HTML]{69BF7B}1  & \cellcolor[HTML]{63BE7B}0  & \cellcolor[HTML]{94CC7D}8  & \cellcolor[HTML]{9BCE7E}9  & \cellcolor[HTML]{88C87D}6  & \cellcolor[HTML]{FA8370}53 & \cellcolor[HTML]{88C87D}6  & \cellcolor[HTML]{F8E983}24 & \cellcolor[HTML]{82C77C}5  & \cellcolor[HTML]{FBA176}45 & \cellcolor[HTML]{FB9D75}46 & \cellcolor[HTML]{F9716D}58 & \cellcolor[HTML]{FB9273}49 & \cellcolor[HTML]{FECA7E}34 & \cellcolor[HTML]{FFE082}28 & \cellcolor[HTML]{FA8370}53 & \cellcolor[HTML]{88C87D}6  & \cellcolor[HTML]{82C77C}5  & \cellcolor[HTML]{82C77C}5  \\
\textbf{fr}      & \cellcolor[HTML]{6FC17B}2  & \cellcolor[HTML]{8ECA7D}7  & \cellcolor[HTML]{94CC7D}8  & \cellcolor[HTML]{63BE7B}0  & \cellcolor[HTML]{6FC17B}2  & \cellcolor[HTML]{7BC57C}4  & \cellcolor[HTML]{FA8370}53 & \cellcolor[HTML]{ADD37F}12 & \cellcolor[HTML]{FFE884}26 & \cellcolor[HTML]{A1D07E}10 & \cellcolor[HTML]{FB9D75}46 & \cellcolor[HTML]{FB9A75}47 & \cellcolor[HTML]{F96D6C}59 & \cellcolor[HTML]{FA8B72}51 & \cellcolor[HTML]{FDC37D}36 & \cellcolor[HTML]{FED981}30 & \cellcolor[HTML]{FA8370}53 & \cellcolor[HTML]{B4D57F}13 & \cellcolor[HTML]{ADD37F}12 & \cellcolor[HTML]{ADD37F}12 \\
\textbf{es}      & \cellcolor[HTML]{69BF7B}1  & \cellcolor[HTML]{94CC7D}8  & \cellcolor[HTML]{9BCE7E}9  & \cellcolor[HTML]{6FC17B}2  & \cellcolor[HTML]{63BE7B}0  & \cellcolor[HTML]{75C37C}3  & \cellcolor[HTML]{FA8070}54 & \cellcolor[HTML]{B4D57F}13 & \cellcolor[HTML]{FFE483}27 & \cellcolor[HTML]{A7D17E}11 & \cellcolor[HTML]{FB9D75}46 & \cellcolor[HTML]{FB9A75}47 & \cellcolor[HTML]{F96D6C}59 & \cellcolor[HTML]{FA8B72}51 & \cellcolor[HTML]{FDBF7C}37 & \cellcolor[HTML]{FED580}31 & \cellcolor[HTML]{FA8070}54 & \cellcolor[HTML]{BAD780}14 & \cellcolor[HTML]{ADD37F}12 & \cellcolor[HTML]{B4D57F}13 \\
\textbf{pt}      & \cellcolor[HTML]{7BC57C}4  & \cellcolor[HTML]{82C77C}5  & \cellcolor[HTML]{88C87D}6  & \cellcolor[HTML]{7BC57C}4  & \cellcolor[HTML]{75C37C}3  & \cellcolor[HTML]{63BE7B}0  & \cellcolor[HTML]{FA8370}53 & \cellcolor[HTML]{A7D17E}11 & \cellcolor[HTML]{FFE884}26 & \cellcolor[HTML]{9BCE7E}9  & \cellcolor[HTML]{FBA176}45 & \cellcolor[HTML]{FB9D75}46 & \cellcolor[HTML]{F96D6C}59 & \cellcolor[HTML]{FA8F73}50 & \cellcolor[HTML]{FDC37D}36 & \cellcolor[HTML]{FFDD82}29 & \cellcolor[HTML]{FA8370}53 & \cellcolor[HTML]{A7D17E}11 & \cellcolor[HTML]{9BCE7E}9  & \cellcolor[HTML]{9BCE7E}9  \\
\textbf{ar}      & \cellcolor[HTML]{FA8070}54 & \cellcolor[HTML]{FA8370}53 & \cellcolor[HTML]{FA8370}53 & \cellcolor[HTML]{FA8370}53 & \cellcolor[HTML]{FA8070}54 & \cellcolor[HTML]{FA8370}53 & \cellcolor[HTML]{63BE7B}0  & \cellcolor[HTML]{FA8F73}50 & \cellcolor[HTML]{FFDD82}29 & \cellcolor[HTML]{FB9273}49 & \cellcolor[HTML]{ADD37F}12 & \cellcolor[HTML]{8ECA7D}7  & \cellcolor[HTML]{9BCE7E}9  & \cellcolor[HTML]{8ECA7D}7  & \cellcolor[HTML]{D9E081}19 & \cellcolor[HTML]{FED981}30 & \cellcolor[HTML]{82C77C}5  & \cellcolor[HTML]{FA8B72}51 & \cellcolor[HTML]{FA8771}52 & \cellcolor[HTML]{FA8771}52 \\
\textbf{vi}      & \cellcolor[HTML]{BAD780}14 & \cellcolor[HTML]{8ECA7D}7  & \cellcolor[HTML]{88C87D}6  & \cellcolor[HTML]{ADD37F}12 & \cellcolor[HTML]{B4D57F}13 & \cellcolor[HTML]{A7D17E}11 & \cellcolor[HTML]{FA8F73}50 & \cellcolor[HTML]{63BE7B}0  & \cellcolor[HTML]{ECE582}22 & \cellcolor[HTML]{82C77C}5  & \cellcolor[HTML]{FCAC78}42 & \cellcolor[HTML]{FCA978}43 & \cellcolor[HTML]{F97C6F}55 & \cellcolor[HTML]{FB9D75}46 & \cellcolor[HTML]{FED580}31 & \cellcolor[HTML]{FFE884}26 & \cellcolor[HTML]{FA8F73}50 & \cellcolor[HTML]{82C77C}5  & \cellcolor[HTML]{88C87D}6  & \cellcolor[HTML]{88C87D}6  \\
\textbf{hi}      & \cellcolor[HTML]{FFE082}28 & \cellcolor[HTML]{F8E983}24 & \cellcolor[HTML]{F8E983}24 & \cellcolor[HTML]{FFE884}26 & \cellcolor[HTML]{FFE483}27 & \cellcolor[HTML]{FFE884}26 & \cellcolor[HTML]{FFDD82}29 & \cellcolor[HTML]{ECE582}22 & \cellcolor[HTML]{63BE7B}0  & \cellcolor[HTML]{DFE282}20 & \cellcolor[HTML]{F2E783}23 & \cellcolor[HTML]{ECE582}22 & \cellcolor[HTML]{FECA7E}34 & \cellcolor[HTML]{FFEB84}25 & \cellcolor[HTML]{A7D17E}11 & \cellcolor[HTML]{9BCE7E}9  & \cellcolor[HTML]{FFDD82}29 & \cellcolor[HTML]{F2E783}23 & \cellcolor[HTML]{F2E783}23 & \cellcolor[HTML]{F8E983}24 \\
\textbf{id}      & \cellcolor[HTML]{ADD37F}12 & \cellcolor[HTML]{88C87D}6  & \cellcolor[HTML]{82C77C}5  & \cellcolor[HTML]{A1D07E}10 & \cellcolor[HTML]{A7D17E}11 & \cellcolor[HTML]{9BCE7E}9  & \cellcolor[HTML]{FB9273}49 & \cellcolor[HTML]{82C77C}5  & \cellcolor[HTML]{DFE282}20 & \cellcolor[HTML]{63BE7B}0  & \cellcolor[HTML]{FCB079}41 & \cellcolor[HTML]{FCAC78}42 & \cellcolor[HTML]{FA8070}54 & \cellcolor[HTML]{FBA176}45 & \cellcolor[HTML]{FED981}30 & \cellcolor[HTML]{F2E783}23 & \cellcolor[HTML]{FB9273}49 & \cellcolor[HTML]{88C87D}6  & \cellcolor[HTML]{88C87D}6  & \cellcolor[HTML]{82C77C}5  \\
\textbf{bn}      & \cellcolor[HTML]{FB9A75}47 & \cellcolor[HTML]{FBA176}45 & \cellcolor[HTML]{FBA176}45 & \cellcolor[HTML]{FB9D75}46 & \cellcolor[HTML]{FB9D75}46 & \cellcolor[HTML]{FBA176}45 & \cellcolor[HTML]{ADD37F}12 & \cellcolor[HTML]{FCAC78}42 & \cellcolor[HTML]{F2E783}23 & \cellcolor[HTML]{FCB079}41 & \cellcolor[HTML]{63BE7B}0  & \cellcolor[HTML]{A1D07E}10 & \cellcolor[HTML]{DFE282}20 & \cellcolor[HTML]{9BCE7E}9  & \cellcolor[HTML]{BAD780}14 & \cellcolor[HTML]{FFE884}26 & \cellcolor[HTML]{9BCE7E}9  & \cellcolor[HTML]{FCA978}43 & \cellcolor[HTML]{FCA978}43 & \cellcolor[HTML]{FCA577}44 \\
\textbf{ta}      & \cellcolor[HTML]{FB9674}48 & \cellcolor[HTML]{FB9D75}46 & \cellcolor[HTML]{FB9D75}46 & \cellcolor[HTML]{FB9A75}47 & \cellcolor[HTML]{FB9A75}47 & \cellcolor[HTML]{FB9D75}46 & \cellcolor[HTML]{8ECA7D}7  & \cellcolor[HTML]{FCA978}43 & \cellcolor[HTML]{ECE582}22 & \cellcolor[HTML]{FCAC78}42 & \cellcolor[HTML]{A1D07E}10 & \cellcolor[HTML]{63BE7B}0  & \cellcolor[HTML]{B4D57F}13 & \cellcolor[HTML]{88C87D}6  & \cellcolor[HTML]{B4D57F}13 & \cellcolor[HTML]{ECE582}22 & \cellcolor[HTML]{9BCE7E}9  & \cellcolor[HTML]{FCA577}44 & \cellcolor[HTML]{FBA176}45 & \cellcolor[HTML]{FBA176}45 \\
\textbf{te}      & \cellcolor[HTML]{F8696B}60 & \cellcolor[HTML]{F9716D}58 & \cellcolor[HTML]{F9716D}58 & \cellcolor[HTML]{F96D6C}59 & \cellcolor[HTML]{F96D6C}59 & \cellcolor[HTML]{F96D6C}59 & \cellcolor[HTML]{9BCE7E}9  & \cellcolor[HTML]{F97C6F}55 & \cellcolor[HTML]{FECA7E}34 & \cellcolor[HTML]{FA8070}54 & \cellcolor[HTML]{DFE282}20 & \cellcolor[HTML]{B4D57F}13 & \cellcolor[HTML]{63BE7B}0  & \cellcolor[HTML]{B4D57F}13 & \cellcolor[HTML]{FFEB84}25 & \cellcolor[HTML]{FED17F}32 & \cellcolor[HTML]{ADD37F}12 & \cellcolor[HTML]{F9786E}56 & \cellcolor[HTML]{F9756E}57 & \cellcolor[HTML]{F9756E}57 \\
\textbf{ur}      & \cellcolor[HTML]{FA8771}52 & \cellcolor[HTML]{FB9273}49 & \cellcolor[HTML]{FB9273}49 & \cellcolor[HTML]{FA8B72}51 & \cellcolor[HTML]{FA8B72}51 & \cellcolor[HTML]{FA8F73}50 & \cellcolor[HTML]{8ECA7D}7  & \cellcolor[HTML]{FB9D75}46 & \cellcolor[HTML]{FFEB84}25 & \cellcolor[HTML]{FBA176}45 & \cellcolor[HTML]{9BCE7E}9  & \cellcolor[HTML]{88C87D}6  & \cellcolor[HTML]{B4D57F}13 & \cellcolor[HTML]{63BE7B}0  & \cellcolor[HTML]{C0D980}15 & \cellcolor[HTML]{FFE884}26 & \cellcolor[HTML]{88C87D}6  & \cellcolor[HTML]{FB9A75}47 & \cellcolor[HTML]{FB9674}48 & \cellcolor[HTML]{FB9674}48 \\
\textbf{ne}      & \cellcolor[HTML]{FDBB7B}38 & \cellcolor[HTML]{FECA7E}34 & \cellcolor[HTML]{FECA7E}34 & \cellcolor[HTML]{FDC37D}36 & \cellcolor[HTML]{FDBF7C}37 & \cellcolor[HTML]{FDC37D}36 & \cellcolor[HTML]{D9E081}19 & \cellcolor[HTML]{FED580}31 & \cellcolor[HTML]{A7D17E}11 & \cellcolor[HTML]{FED981}30 & \cellcolor[HTML]{BAD780}14 & \cellcolor[HTML]{B4D57F}13 & \cellcolor[HTML]{FFEB84}25 & \cellcolor[HTML]{C0D980}15 & \cellcolor[HTML]{63BE7B}0  & \cellcolor[HTML]{B4D57F}13 & \cellcolor[HTML]{DFE282}20 & \cellcolor[HTML]{FED17F}32 & \cellcolor[HTML]{FECE7F}33 & \cellcolor[HTML]{FECE7F}33 \\
\textbf{mr}      & \cellcolor[HTML]{FED580}31 & \cellcolor[HTML]{FFE082}28 & \cellcolor[HTML]{FFE082}28 & \cellcolor[HTML]{FED981}30 & \cellcolor[HTML]{FED580}31 & \cellcolor[HTML]{FFDD82}29 & \cellcolor[HTML]{FED981}30 & \cellcolor[HTML]{FFE884}26 & \cellcolor[HTML]{9BCE7E}9  & \cellcolor[HTML]{F2E783}23 & \cellcolor[HTML]{FFE884}26 & \cellcolor[HTML]{ECE582}22 & \cellcolor[HTML]{FED17F}32 & \cellcolor[HTML]{FFE884}26 & \cellcolor[HTML]{B4D57F}13 & \cellcolor[HTML]{63BE7B}0  & \cellcolor[HTML]{FED580}31 & \cellcolor[HTML]{FFE884}26 & \cellcolor[HTML]{FFE884}26 & \cellcolor[HTML]{FFE884}26 \\
\textbf{gu}      & \cellcolor[HTML]{F97C6F}55 & \cellcolor[HTML]{FA8370}53 & \cellcolor[HTML]{FA8370}53 & \cellcolor[HTML]{FA8370}53 & \cellcolor[HTML]{FA8070}54 & \cellcolor[HTML]{FA8370}53 & \cellcolor[HTML]{82C77C}5  & \cellcolor[HTML]{FA8F73}50 & \cellcolor[HTML]{FFDD82}29 & \cellcolor[HTML]{FB9273}49 & \cellcolor[HTML]{9BCE7E}9  & \cellcolor[HTML]{9BCE7E}9  & \cellcolor[HTML]{ADD37F}12 & \cellcolor[HTML]{88C87D}6  & \cellcolor[HTML]{DFE282}20 & \cellcolor[HTML]{FED580}31 & \cellcolor[HTML]{63BE7B}0  & \cellcolor[HTML]{FA8B72}51 & \cellcolor[HTML]{FA8B72}51 & \cellcolor[HTML]{FA8771}52 \\
\textbf{sw}      & \cellcolor[HTML]{C0D980}15 & \cellcolor[HTML]{8ECA7D}7  & \cellcolor[HTML]{88C87D}6  & \cellcolor[HTML]{B4D57F}13 & \cellcolor[HTML]{BAD780}14 & \cellcolor[HTML]{A7D17E}11 & \cellcolor[HTML]{FA8B72}51 & \cellcolor[HTML]{82C77C}5  & \cellcolor[HTML]{F2E783}23 & \cellcolor[HTML]{88C87D}6  & \cellcolor[HTML]{FCA978}43 & \cellcolor[HTML]{FCA577}44 & \cellcolor[HTML]{F9786E}56 & \cellcolor[HTML]{FB9A75}47 & \cellcolor[HTML]{FED17F}32 & \cellcolor[HTML]{FFE884}26 & \cellcolor[HTML]{FA8B72}51 & \cellcolor[HTML]{63BE7B}0  & \cellcolor[HTML]{82C77C}5  & \cellcolor[HTML]{7BC57C}4  \\
\textbf{yo}      & \cellcolor[HTML]{B4D57F}13 & \cellcolor[HTML]{82C77C}5  & \cellcolor[HTML]{82C77C}5  & \cellcolor[HTML]{ADD37F}12 & \cellcolor[HTML]{ADD37F}12 & \cellcolor[HTML]{9BCE7E}9  & \cellcolor[HTML]{FA8771}52 & \cellcolor[HTML]{88C87D}6  & \cellcolor[HTML]{F2E783}23 & \cellcolor[HTML]{88C87D}6  & \cellcolor[HTML]{FCA978}43 & \cellcolor[HTML]{FBA176}45 & \cellcolor[HTML]{F9756E}57 & \cellcolor[HTML]{FB9674}48 & \cellcolor[HTML]{FECE7F}33 & \cellcolor[HTML]{FFE884}26 & \cellcolor[HTML]{FA8B72}51 & \cellcolor[HTML]{82C77C}5  & \cellcolor[HTML]{63BE7B}0  & \cellcolor[HTML]{75C37C}3  \\
\textbf{ig}      & \cellcolor[HTML]{B4D57F}13 & \cellcolor[HTML]{82C77C}5  & \cellcolor[HTML]{82C77C}5  & \cellcolor[HTML]{ADD37F}12 & \cellcolor[HTML]{B4D57F}13 & \cellcolor[HTML]{9BCE7E}9  & \cellcolor[HTML]{FA8771}52 & \cellcolor[HTML]{88C87D}6  & \cellcolor[HTML]{F8E983}24 & \cellcolor[HTML]{82C77C}5  & \cellcolor[HTML]{FCA577}44 & \cellcolor[HTML]{FBA176}45 & \cellcolor[HTML]{F9756E}57 & \cellcolor[HTML]{FB9674}48 & \cellcolor[HTML]{FECE7F}33 & \cellcolor[HTML]{FFE884}26 & \cellcolor[HTML]{FA8771}52 & \cellcolor[HTML]{7BC57C}4  & \cellcolor[HTML]{75C37C}3  & \cellcolor[HTML]{63BE7B}0  \\
\textbf{Average} & 25.2                       & 22.45                      & 22.5                       & 24.15                      & 24.6                       & 23.4                       & 34.6                       & 22.5                       & 22.45                      & 21.4                       & 30.5                       & 30.05                      & 39.5                       & 32.15                      & 25.2                       & 24.65                      & 34.75                      & 22.95                      & 22.7                       & 22.8                      
\end{tabular}}
\caption{Original distance matrix generated from the BLOOM-560m model.}
\label{dist560m}

\end{table*}

\begin{appendix}
\end{appendix}

\end{document}